\documentclass[lettersize,journal]{IEEEtran}
\usepackage{amsmath,amsfonts}
\usepackage{algorithmic}
\usepackage{algorithm}
\usepackage{array}
\usepackage[caption=false,font=normalsize,labelfont=sf,textfont=sf]{subfig}
\usepackage{textcomp}
\usepackage{stfloats}
\usepackage{url}
\usepackage{verbatim}
\usepackage{graphicx}
\usepackage{subcaption}
\usepackage{booktabs}
\usepackage{xcolor}
\usepackage[colorlinks=true, linkcolor=blue, citecolor=blue, urlcolor=blue]{hyperref}

\usepackage{caption}
\usepackage{cite}
\begin{document}

\title{CDPR: Cross-modal Diffusion with Polarization for\\ Reliable Monocular Depth Estimation}

\author{Rongjia Yu, Tong Jia, Hao Wang, Xiaofang Li, Xiao Yang, Zinuo Zhang, Cuiwei Liu
\thanks{This work was supported in part by the National Key Research and Development Project of China under Grant 2022YFF0902401, in part by the National Natural Science Foundation of China under Grant U22A2063 and 62173083, in part by the China Postdoctoral Science Foundation under No. 2024T170114, in part by the Major Program of National Natural Science Foundation of China under Grant 71790614, in part by the 111 Project under Grant B16009, and in part by the Liaoning Provincial "Selecting the Best Candidates by Opening Competition Mechanism" Science and Technology Program under Grant 2023JH1/10400045; the Fundamental Research Funds for the Central Universities under Grant N2424022.}
}



\maketitle

\begin{abstract}
Monocular depth estimation is a fundamental yet challenging task in computer vision, especially under complex conditions such as textureless surfaces, transparency, and specular reflections. Recent diffusion-based approaches have significantly advanced performance by reformulating depth prediction as a denoising process in the latent space. However, existing methods rely solely on RGB inputs, which often lack sufficient cues in challenging regions. In this work, we present CDPR — Cross-modal Diffusion with Polarization for Reliable Monocular Depth Estimation — a novel diffusion-based framework that integrates physically grounded polarization priors to enhance estimation robustness. Specifically, we encode both RGB and polarization (AoLP/DoLP) images into a shared latent space via a pre-trained Variational Autoencoder (VAE), and dynamically fuse multi-modal information through a learnable confidence-aware gating mechanism. This fusion module adaptively suppresses noisy signals in polarization inputs while preserving informative cues, particularly around reflective or transparent surfaces, and provides the integrated latent representation for subsequent monocular depth estimation. Beyond depth estimation, we further verify that our framework can be easily generalized to surface normal prediction with minimal modification, showcasing its scalability to general polarization-guided dense prediction tasks. Experiments on both synthetic and real-world datasets validate that CDPR significantly outperforms RGB-only baselines in challenging regions while maintaining competitive performance in standard scenes. The source code of CDPR is available at \href{https://github.com/YrjNEU/CDPR-main.git}{https://github.com/YrjNEU/CDPR-main.git}.

\end{abstract}


\section{Introduction}
\IEEEPARstart{M}{onocular} depth estimation is a fundamental yet challenging task in computer vision, supporting numerous downstream applications such as image segmentation \cite{park2017rdfnet}, robotic grasp detection \cite{lenz2015deep}, and human pose recognition \cite{shotton2011real}. Recent years have witnessed the success of discriminative models driven by large-scale datasets and powerful neural architectures. Techniques such as dataset expansion\cite{yin2023metric3d}\cite{yang2024depth}\cite{yang2024depthv2}\cite{wang2025moge}\cite{wang2025moge2}, unsupervised learning strategies\cite{ling2021unsupervised}\cite{peng2021excavating}\cite{li2022self}\cite{sun2023sc}\cite{yu2024vfm} and self-supervised learning strategies\cite{zhou2017unsupervised}\cite{mahjourian2018unsupervised}\cite{shao2024monodiffusion} have substantially improved model generalization and prediction accuracy.

However, these discriminative approaches face a critical trade-off: models with strong generalization ability typically require a massive amount of labeled training data, while those with fewer data demands often suffer from performance degradation when applied to unseen scenes. To alleviate this dependency, generative paradigms have gained attention for their superior structural reasoning and data efficiency\cite{cs2018monocular}\cite{aleotti2018generative}. In particular, diffusion-based models \cite{shao2024monodiffusion}\cite{ke2024repurposing}\cite{fu2024geowizard} have demonstrated promising results by reformulating depth estimation as a denoising process in latent space. By transferring priors from image generation tasks, these models enhance global scene understanding and show better performance under limited supervision. Recent works \cite{garcia2025fine}\cite{gui2025depthfm}\cite{he2024lotus} have also explored fast inference strategies by reducing denoising steps. Despite these advances, most existing diffusion models are constrained by RGB-only inputs, which often fail to provide sufficient cues in ambiguous regions such as textureless areas, transparent surfaces, or specular reflections.

\begin{figure}[!t]
\centering
\subfloat[{\scriptsize \rmfamily GT Point Cloud}]{
    \includegraphics[scale=0.06]{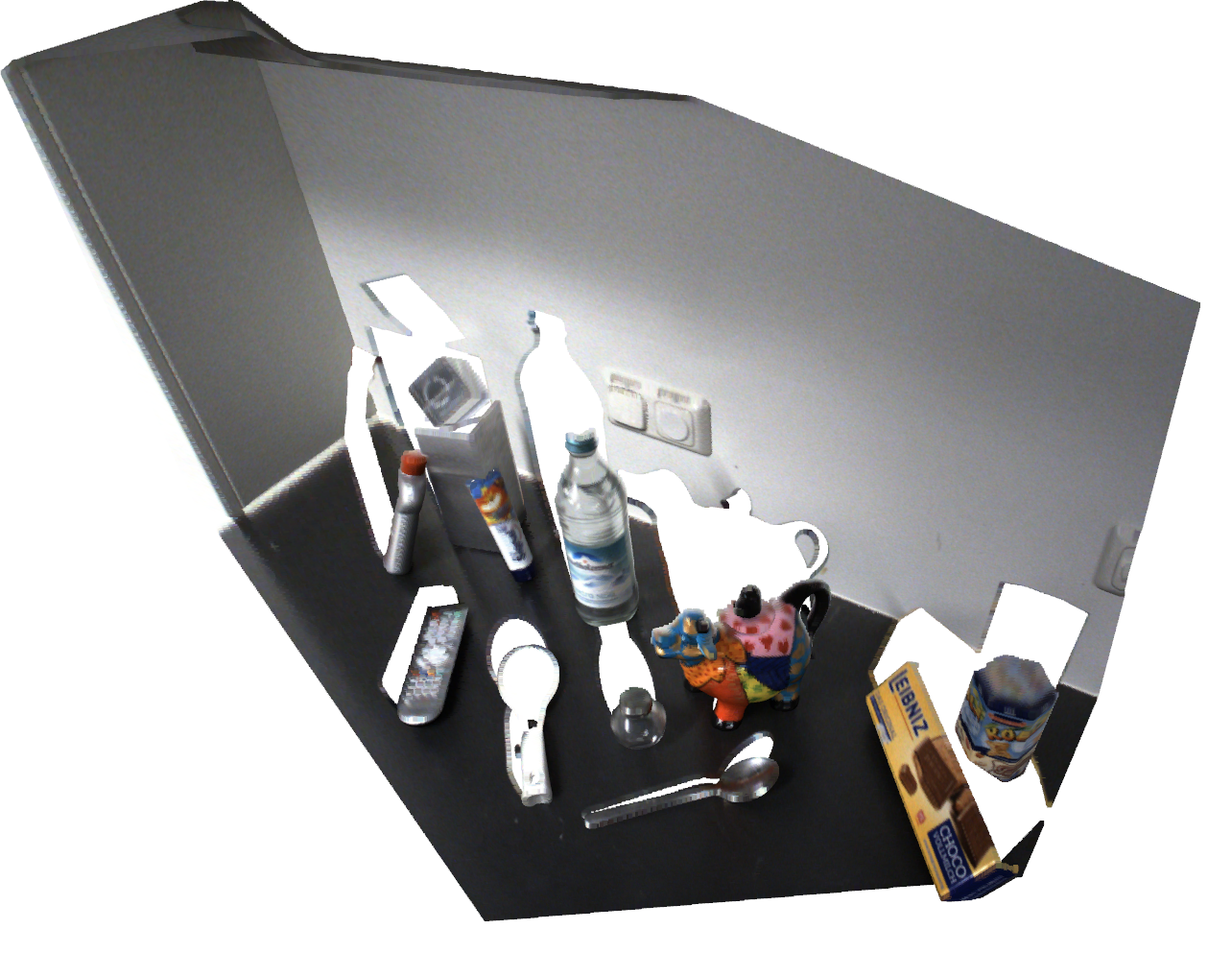}}
\hspace{0.1mm}
\subfloat[{\scriptsize \rmfamily Marigold Point Cloud}]{
    \includegraphics[scale=0.06]{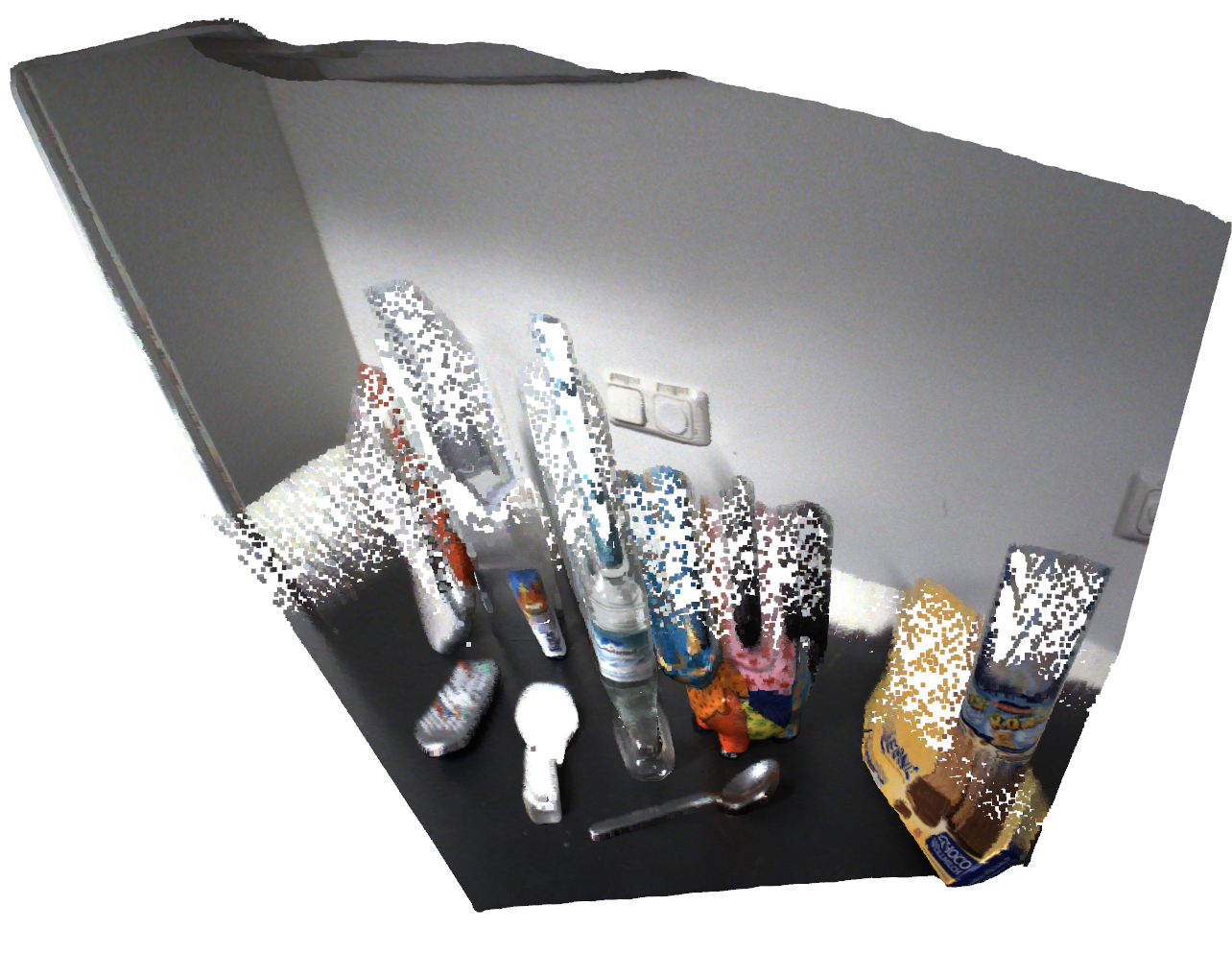}}
\hspace{0.1mm}
\subfloat[{\scriptsize \rmfamily Ours Point Cloud}]{
    \includegraphics[scale=0.06]{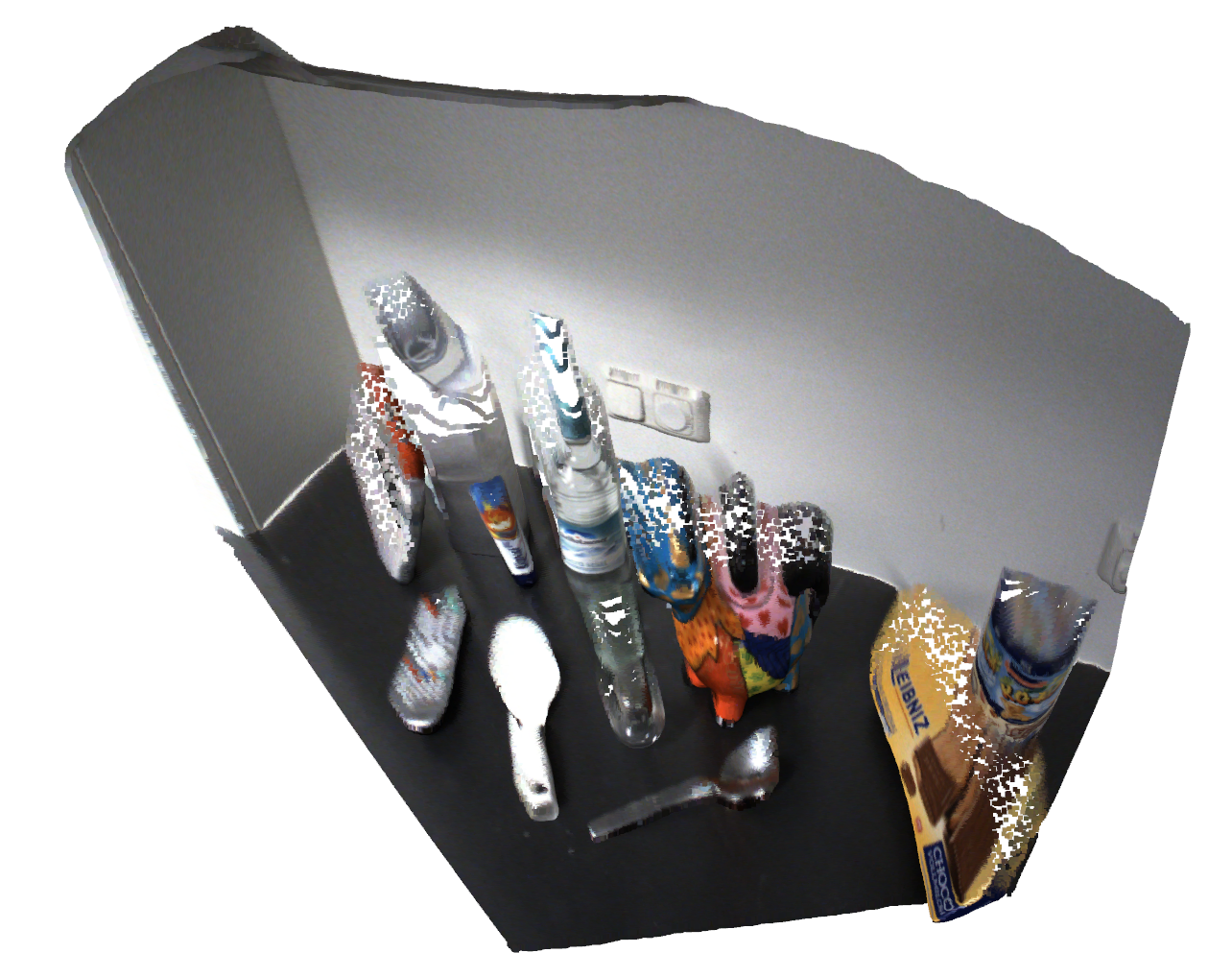}}

\vspace{1mm} 

\subfloat[{\scriptsize \rmfamily GT Depth}]{
    \includegraphics[scale=0.065]{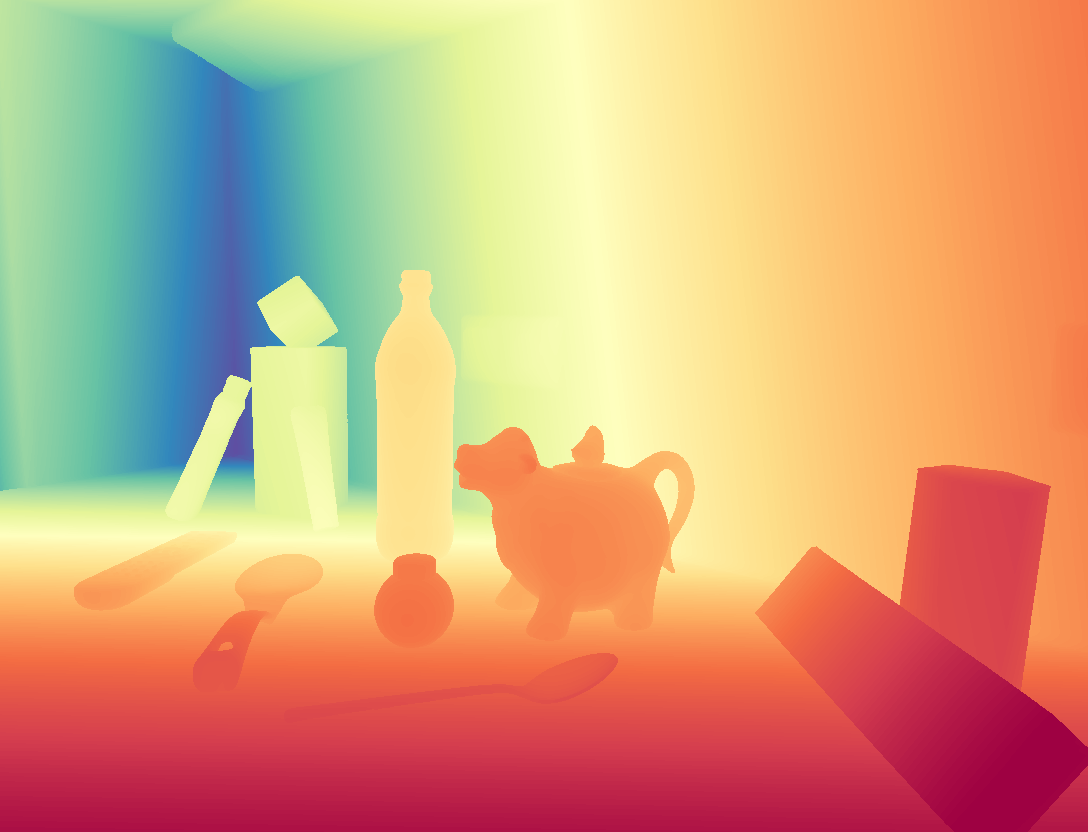}}
\hspace{0.5mm}
\subfloat[{\scriptsize \rmfamily Marigold Depth}]{
    \includegraphics[scale=0.065]{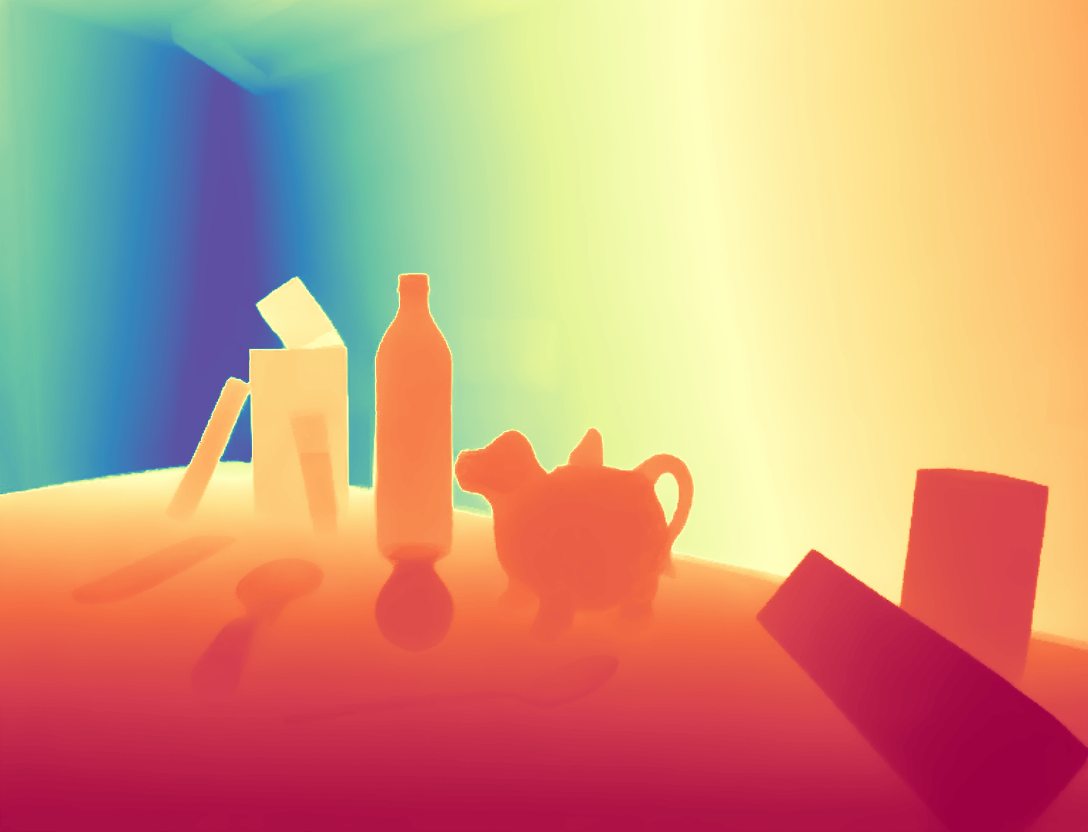}}
\hspace{0.5mm}
\subfloat[{\scriptsize \rmfamily Ours Depth}]{
    \includegraphics[scale=0.065]{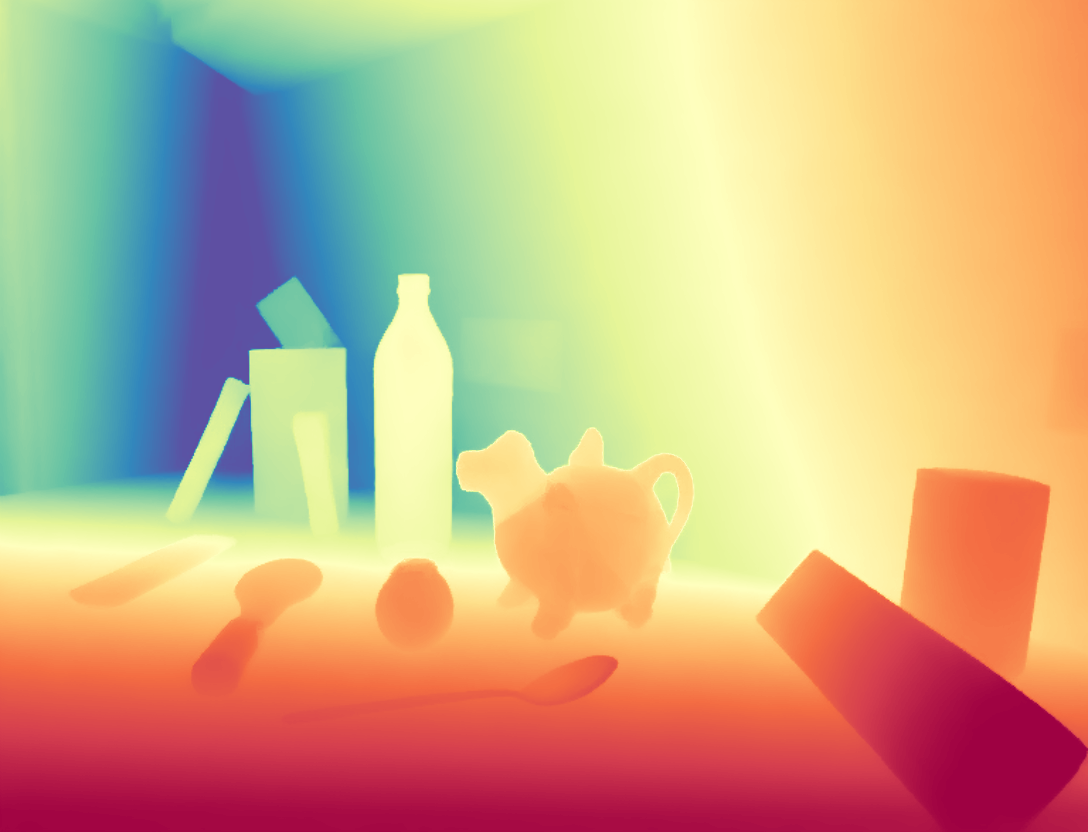}}

\vspace{1mm} 

\subfloat[{\scriptsize \rmfamily RGB}]{
    \includegraphics[scale=0.065]{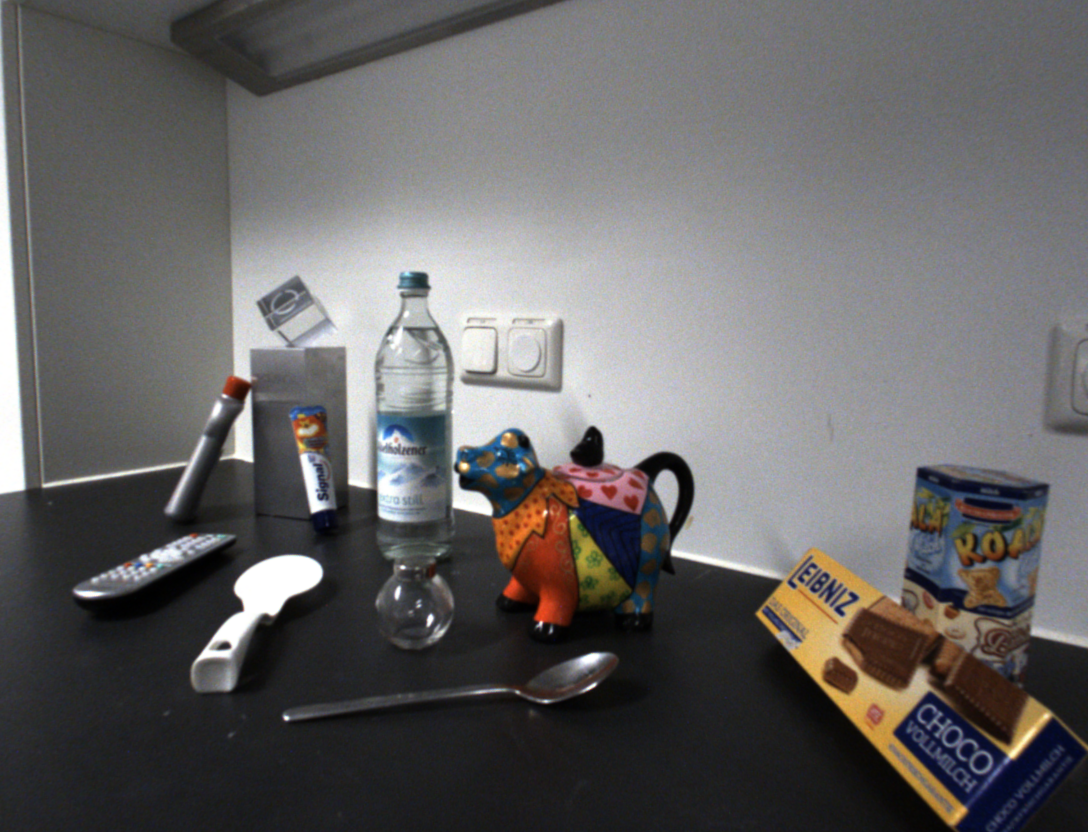}}
\hspace{0.5mm}
\subfloat[{\scriptsize \rmfamily AoLP}]{
    \includegraphics[scale=0.065]{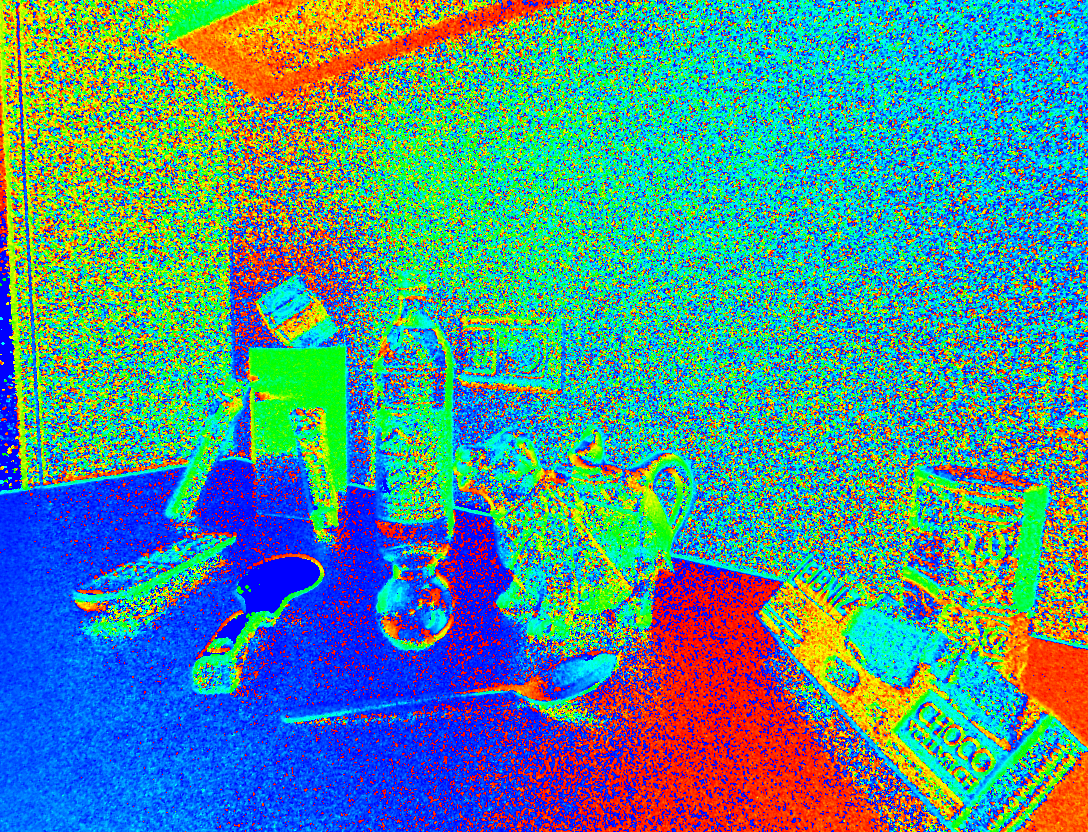}}
\hspace{0.5mm}
\subfloat[{\scriptsize \rmfamily DoLP}]{
    \includegraphics[scale=0.065]{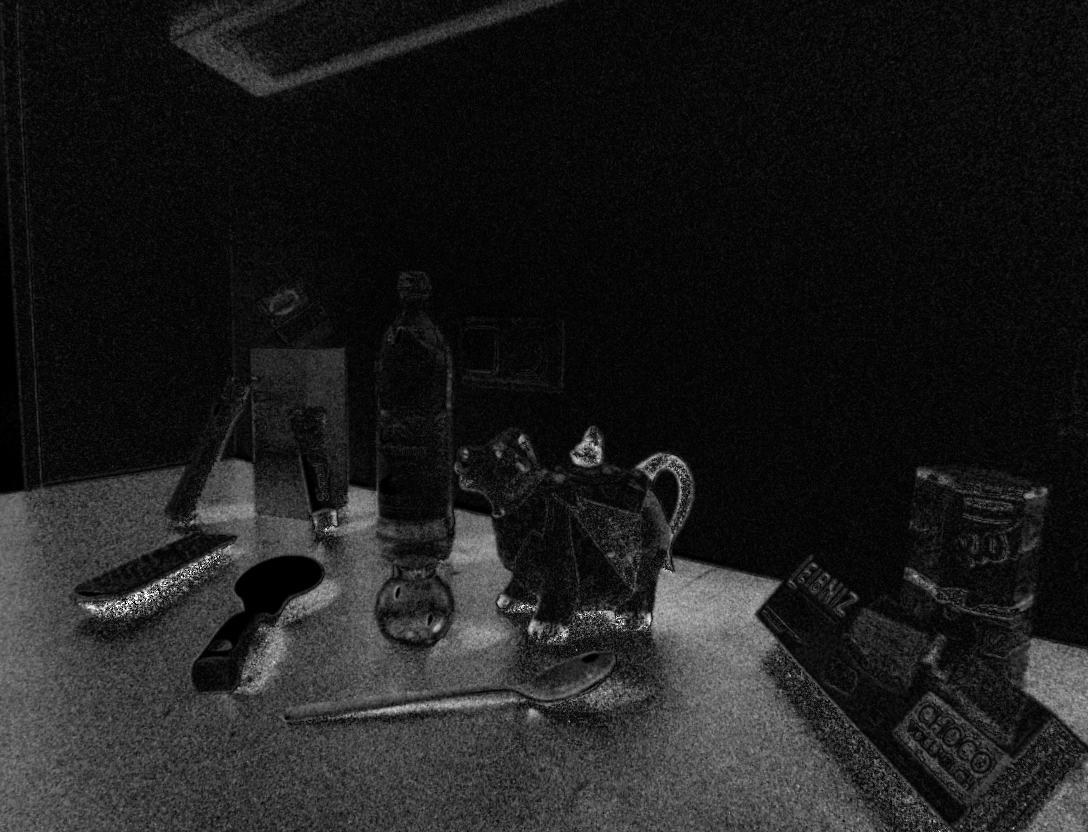}}
\caption{\rmfamily RGB Methods (e.g. Marigold\cite{ke2024repurposing}) suffers from scale ambiguity in low-texture regions, such as the table surface, where the predicted distance to the background wall is significantly overestimated. While this error may have limited impact on perceptual quality even standard metrics, it is substantially amplified in downstream tasks like 3D reconstruction, leading to distorted geometry and missing surfaces. In contrast, our polarization-guided fusion approach mitigates these issues effectively, producing more accurate depth predictions under challenging visual conditions. The resulting point clouds are geometrically more consistent and complete, demonstrating the strong potential of polarization cues in enhancing robust depth estimation.}
\label{fig_1}
\end{figure}

To improve depth estimation in these challenging scenarios, we incorporate physical cues from polarization images, which encode rich surface geometry and material properties that are especially useful in optically complex regions, e.g. glass or metal surfaces \cite{ba2020deep}, and also mitigate scale ambiguity in textureless environments. As shown in Fig.~\ref{fig_1}, RGB-only methods (e.g. Marigold \cite{ke2024repurposing}) severely overestimate background depth, while our polarization-guided framework recovers more accurate and coherent geometry.

Nevertheless, two major challenges hinder the use of polarization in dense prediction. First, the lack of large-scale polarization datasets with ground truth depth restricts effective training and benchmarking. Although diffusion models ease the overall data burden, fine-tuning a strong generative model still requires tens of thousands of labeled samples to ensure robustness and generalization. Second, polarization signals are inherently noisy and sparse, making naive fusion with RGB inputs suboptimal.

To address these issues, we propose CDPR, a Cross-modal Diffusion framework with Polarization for Reliable Monocular Depth Estimation. Specifically, we introduce a confidence-aware fusion module that dynamically integrates RGB and polarization features in the latent space via learnable latent confidence maps, enhancing complementary information flow. Additionally, based on Hypersim dataset\cite{roberts2021hypersim}, we construct a synthetic dataset, HyperPol, with physically grounded polarization and depth annotations to support scalable training.  It is important to note that both RGB and polarization cues are captured simultaneously from a single-shot acquisition of a single polarization camera, ensuring that our framework remains strictly monocular while leveraging richer physical information.

Experimental results demonstrate that our method has a satisfactory performance across multiple benchmarks, especially exhibiting superior structural consistency and prediction stability in transparent and reflective regions. In addition to depth estimation, our framework can be directly extended to surface normal prediction, showing comparable robustness and accuracy, which further verifies its general applicability to polarization-guided dense prediction tasks.

Our contributions can be summarized as follows:

1. We propose a diffusion-based, confidence-guided fusion framework that effectively incorporates physically grounded polarization cues for monocular depth estimation, enhancing depth recovery in optically complex regions. The framework is further extended to surface normal estimation, demonstrating its generality across polarization-guided dense prediction tasks;

2. We design a lightweight confidence predictor to generate spatially adaptive $\alpha$-maps for region-wise fusion of RGB and polarization features in latent space, enabling dynamic cross-modal information flow;

3. We construct a synthetic polarization dataset HyperPol, which generates physically consistent AoLP and DoLP images based on the Hypersim dataset, providing scalable training resources for polarization-assisted depth estimation.

The remainder of this paper is organized as follows:

In Section II, we review related works on RGB-based monocular depth estimation and polarization-guided vision tasks. Section III provides the preliminary concepts necessary to understand our proposed framework. In Section IV, we describe the detailed implementation of CDPR, including the synthetic dataset construction, polarization encoding, confidence prediction module, gated latent fusion strategy, and depth estimation method in latent space. Section V presents the experimental setup, followed by comprehensive evaluations on both depth and surface normal estimation tasks. We also conduct ablation studies to analyze the contribution of each component. Finally, we conclude the paper in Section VI by summarizing our contributions, discussing current limitations, and outlining potential directions for future research.

\section{Related Works}
\subsection{RGB-Based Monocular Depth Estimation}
Depth estimation is a fundamental component of 3D scene understanding. RGB-based monocular depth estimation has long been a core problem in computer vision. With the rapid advancement of computational hardware, deep learning-based approaches have become the dominant direction in this field. Eigen et al. \cite{eigen2014depth} first proposed an end-to-end convolutional neural network for monocular depth estimation, inspiring numerous follow-up works focusing on improving model accuracy \cite{yin2023metric3d}\cite{yang2024depth}\cite{yang2024depthv2}\cite{ke2024repurposing}\cite{ranftl2021vision}\cite{ranftl2020towards}\cite{song2021monocular} and extending task generality (e.g. by jointly estimating surface normals \cite{wang2025moge}\cite{wang2025moge2}\cite{fu2024geowizard}\cite{garcia2025fine}\cite{he2024lotus}\cite{xu2024matters}\cite{eftekhar2021omnidata}).

In recent years, adapting pretrained generative models to dense prediction tasks has become an emerging trend. Ke et al. proposed Marigold\cite{ke2024repurposing}, which reformulates monocular depth estimation as a diffusion-based denoising process, significantly boosting performance. GeoWizard\cite{fu2024geowizard} extended this framework to also predict surface normals, enhancing the model’s ability to capture fine geometric details. DepthFM\cite{gui2025depthfm} introduced flow matching into the diffusion pipeline, greatly reducing the number of inference steps with almost no loss of accuracy. GenPercept\cite{xu2024matters} proposed a single-step generation strategy to mitigate surface texture interference during depth prediction. Subsequently, He et al. presented a systematic analysis of diffusion formulations and proposed Lotus\cite{he2024lotus}, a single-step diffusion framework tailored for RGB dense prediction, achieving substantial improvements in both efficiency and accuracy. E2E-FT\cite{garcia2025fine} revealed that inefficiencies in current dense prediction pipelines stem from design flaws in the inference process, and proposed an efficient finetuning strategy to improve performance and speed for both depth and normal estimation.

However,these methods are all limited to dense prediction from a single RGB image, and struggle in physically degraded scenes such as transparent materials, specular surfaces, or textureless regions. Unlike prior works, we introduce polarization information into a diffusion-based monocular depth estimation pipeline for the first time. Polarization images provide rich cues on surface normal direction and material properties through AoLP (Angle of Linear Polarization) and DoLP (Degree of Linear Polarization), offering unique advantages in regions where traditional RGB cues are ambiguous or unreliable.

\subsection{Polarization-Guided Vision Tasks}
In recent years, a growing number of vision tasks have incorporated polarization cues, as surface geometry intrinsically influences the polarization state of reflected light\cite{collett2005field}\cite{miyazaki2003polarization}\cite{atkinson2017polarisation}. This physical correlation serves as a valuable prior, particularly for enhancing depth-related reasoning in complex scenes.

In the domain of 3D object reconstruction, since the observed polarization is jointly influenced by surface geometry, reflectance properties, and lighting conditions, early Shape from Polarization (SfP) approaches typically rely on strong priors regarding reflection models to simplify the inherently ill-posed nature of the problem\cite{rahmann2001reconstruction}. Later, Baek et al. proposed polarization guided SVBRDF\cite{baek2018simultaneous}, which jointly optimizes appearance parameters, surface normals, and the index of refraction. PANDORA\cite{dave2022pandora} represents the first inverse rendering framework that combines polarization images with neural radiance fields (NeRF)\cite{mildenhall2021nerf}. It utilizes coordinate-based multilayer perceptrons (MLPs) to regress surface normals, diffuse radiance, and specular radiance, which are then composed using a simplified renderer to produce the outgoing Stokes vector. Building upon this, NeRSP\cite{han2024nersp} leverages both photometric\cite{dave2022pandora} and geometric\cite{cao2023multi} cues from polarization data to achieve more accurate inverse rendering. Shao et al. introduced TransPIR\cite{shao2024polarimetric}, a polarization-guided inverse rendering method tailored for transparent objects. This was followed by TransSFP\cite{shao2023transparent}, which models polarization confidence to modulate the influence of physics-based priors in surface normal estimation, enabling robust single-view reconstruction. NeISF\cite{li2024neisf} further proposed a multi-view inverse rendering framework that relaxes the assumption of unpolarized illumination, utilizing polarization constraints to reduce normal uncertainty.

For large-scale scene-level recovery, Zhu and Smith \cite{zhu2019depth} proposes a hybrid polarization–RGB stereo system that resolves the inherent normal ambiguities of polarization using an auxiliary RGB viewpoint within a graphical optimization framework. More relevant to monocular settings, P2D \cite{blanchon2021p2d} proposes a video-based self-supervised monocular depth estimation framework, where polarization cues are incorporated and training relies on photometric and polarimetric consistency across consecutive frames in outdoor sequences. Lei et al. presented SPW \cite{lei2022shape}, a data-driven framework guided by physical priors, which estimates dense surface normals from a single polarization image. DPS-Net \cite{tian2023dps} fuses geometric priors with stereo polarization cues to achieve robust disparity estimation in textureless or reflective regions. PPFT \cite{ikemura2024robust}, introduced by Kei et al., demonstrates the utility of polarization guidance in depth completion, particularly in transparent scenarios.

These works collectively illustrate the value of polarization cues for 3D perception. However, they mainly target stereo reconstruction, shape-from-polarization, or video-based self-supervised monocular learning, and do not provide a systematic integration of polarization into a generative monocular depth estimation pipeline. Moreover, existing polarization-guided monocular estimation approaches differ fundamentally from current monocular methods that operate on single-camera, single-capture inputs, as they rely on multi-view cues or temporal consistency during training rather than learning depth from a single capture.

To fill this gap, we propose a diffusion-based monocular depth estimation framework that jointly leverages monocular polarization inputs (AoLP and DoLP) and RGB intensity captured simultaneously in a single shot. Specifically, we introduce a confidence-aware gating mechanism in the latent space, which explicitly models the modality complementarity between RGB and polarization features. This design enables spatially adaptive fusion through learnable confidence maps, enhancing depth prediction accuracy and robustness in challenging regions without introducing substantial computational overhead during training or inference.
\section{Preliminary}
Light is a type of transverse wave, with its vibration direction perpendicular to the propagation direction. When a light wave passes through a polarizer oriented at an angle $\phi_{pol}$ with respect to the horizontal direction, the transmitted intensity is given by:

\begin{equation}
I_{pol} = I_{un}[1 + \rho\cos(2\phi_a - 2\phi_{pol})],
\label{deqn_ex1a}
\end{equation}
where $I_{un}$ denotes the incident light intensity, $\rho$ represents the degree of linear polarization (DoLP),and $\phi_a$ denotes the angle of linear polarization (AoLP).

By rotating a linear polarizer in front of a polarization camera, four images can be captured at different polarizer angles $\phi_{pol}$, yielding intensity measurements $\left\{ I_{0}, I_{45}, I_{90}, I_{135} \right\}$. These values reflect different polarization states of the incoming light and can be modeled as components of a four-element Stokes vector $S = \left\{ S_{0}, S_{1}, S_{2}, S_{3} \right\}$. Assuming that there is no circular polarization is present, Stokes vector $S$ can be expressed as:

\begin{equation}
\label{stokes_equations}
\begin{cases}
S_0 = \dfrac{1}{4}\left(I_0 + I_{45} + I_{90} + I_{135}\right) \\[1em]  
S_1 = \dfrac{1}{2}\left(I_0 - I_{90}\right) \\[1em]  
S_2 = \dfrac{1}{2}\left(I_{45} - I_{135}\right).
\end{cases}
\end{equation}

From the computed Stokes vector, the DoLP and AoLP can be derived as:

\begin{equation}
\label{rho}
\rho = \frac{\sqrt{S_1^2 + S_2^2}}{S_0},
\end{equation}

\begin{equation}
\label{phia}
\phi_a = \frac{1}{2} \arctan\left( \frac{S_2}{S_1} \right).
\end{equation}

Following SPW \cite{lei2022shape}, the polarization angle is known to exhibit two types of ambiguity: $\pi$-ambiguity and diffuse/specular ambiguity. The $\pi$-ambiguity arises from the fact that the value of $\phi_a$ is defined over the range $[0, \pi]$, making it indistinguishable between $\phi_a$ and $\phi_a+\pi$,  as illustrated in Eq.~\ref{deqn_ex1a}. Additionally, depending on whether diffuse or specular reflection dominates, the polarization direction becomes either parallel or perpendicular to the plane of incidence, respectively.

In our framework, the polarization representation is explicitly designed to account for both forms of ambiguity. See more details in Section IV-B.

\begin{figure*}[!t]
\centering
\includegraphics[width=7.3in]{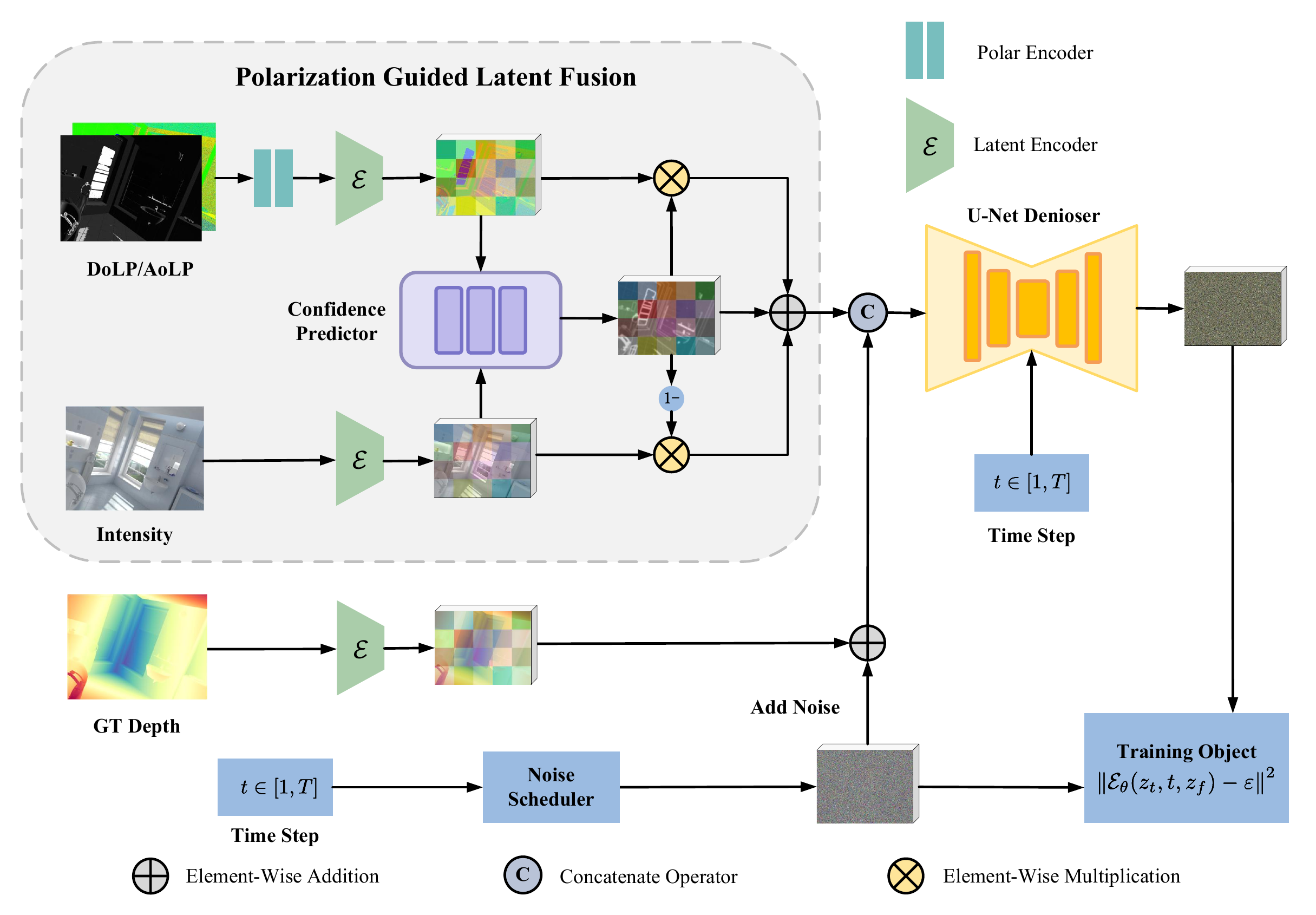}
\caption{Overview of our training pipeline. The input consists of an RGB image and its corresponding polarization observations, i.e., Angle of Linear Polarization (AoLP) and Degree of Linear Polarization (DoLP). The two polarization channels are processed by a Polar Encoder to generate a 3-channel polarization representation. Both RGB and polarization features are separately encoded via VAE encoders. A lightweight confidence predictor then estimates a spatially adaptive $\alpha$-map, which guides gated fusion in the latent space by weighting RGB and polarization features. The fused latent is concatenated with the noisy depth latent and passed to a U-Net denoiser within the diffusion framework. The predicted noise residual is compared with the ground-truth noise to compute the final loss.}
\label{fig_3}
\end{figure*}

\section{Method}
\subsection{Polarization Encoding}
Variational Autoencoder serves as both a latent space compressor and a perceptual aligner between data space and semantic space. Before being fed into the VAE, input data is normalized to the range of $[-1,1]$. While RGB and depth values can be directly rescaled using linear normalization, polarization inputs possess distinct physical meanings and distribution characteristics, necessitating specialized preprocessing.

In this work, we adopt the Angle of Linear Polarization (AoLP) and the Degree of Linear Polarization (DoLP) as input features to represent the polarization modality. 

Specifically, DoLP is a normalized intensity ratio that naturally lies in the range of $[0,1]$. We apply a linear transformation to map it into the range $[-1,1]$ as follows:

\begin{equation}
\boldsymbol{\rho}_{norm} = 2\boldsymbol{\rho}-1.
\end{equation}

AoLP, on the other hand, exhibits inherent periodicity and ambiguity, with a raw value range of $[0,\pi]$, making it unsuitable for direct linear normalization. To address this, we adopt the encoding strategy proposed in SPW \cite{lei2022shape}, where AoLP is transformed into a two-dimensional vector as follows:

\begin{equation}
\boldsymbol{\phi}_{norm} = [\cos 2\boldsymbol{\phi}_a, \sin 2\boldsymbol{\phi}_a].
\end{equation}

This encoding serves two purposes in our task: it eliminates the ambiguity caused by angular periodicity and simultaneously constrains the range of polarization angles to $[-1,1]$, satisfying the input requirements of the VAE. It also compensates for the necessary simplification of $\pi$ and $\pi/2$ ambiguities adopted in our dataset synthesis process.

The final 3-channel polarization input $\boldsymbol{x}_p$ is organized as:

\begin{equation}
\label{eq:phi_vector}
\boldsymbol{x}_p = [\boldsymbol{\rho}_{norm}, \cos 2\boldsymbol{\phi}_a, \sin 2\boldsymbol{\phi}_a].
\end{equation}

Incorporating $\rho_{norm}$ as an auxiliary input helps alleviate the diffuse–specular ambiguity, as regions dominated by specular reflection typically exhibit higher degrees of polarization, providing critical cues for distinguishing surface reflection behaviors.

Although extracting AoLP and DoLP from a polarization camera is computationally simple, ensuring the quality of these measurements in real-world scenarios is substantially more challenging. Polarization cues are highly sensitive to ambient illumination, surface roughness, material reflectance, and sensor noise, often resulting in unstable or unreliable polarization estimates. Such degradation in the input modality motivates the need for a reliability-aware fusion mechanism that can adaptively modulate the influence of polarization features during depth estimation.

\subsection{Confidence Predictor}
Building upon the preprocessing and normalization described in the Section IV-A, the VAE encoders map RGB and polarization inputs into a shared latent space. However, the reliability of the resulting polarization latent features may still vary spatially due to reflectance, material properties, and scene-dependent effects. Treating these latent features as uniformly trustworthy and fusing them unconditionally can introduce semantic imbalance in the latent representation and negatively affect depth estimation. This motivates the need for an adaptive and reliability-aware fusion mechanism.

Inspired by the concept of physics-based polarization confidence introduced in TransSFP \cite{shao2023transparent}, we propose a learnable, lightweight confidence predictor $f_c(\boldsymbol{z}_{RGB},\boldsymbol{z}_{POL})$ that operates in the latent space. This module estimates the reliability of the polarization features at each spatial location. It takes as input the encoded latent features from both RGB and polarization branches and outputs a single-channel confidence map. A Sigmoid activation function is applied at the final layer to constrain the output to the range $[0, 1]$, such that:

\begin{equation}
\label{eq:phi_vector}
\boldsymbol{\alpha} = \sigma[f_c(\boldsymbol{z}_{RGB},\boldsymbol{z}_{POL})].
\end{equation}

Here, $\boldsymbol{\alpha}$ denotes the relative importance of the polarization features at a given spatial location,while $1-\boldsymbol{\alpha}$ corresponds to the complementary weight assigned to the RGB features. The confidence predictor serves as the foundational module for the fusion mechanism, with its output later utilized in the latent-space gating strategy described in Section IV-D.

\begin{figure}[!t]
\centering
\includegraphics[width=3.5in]{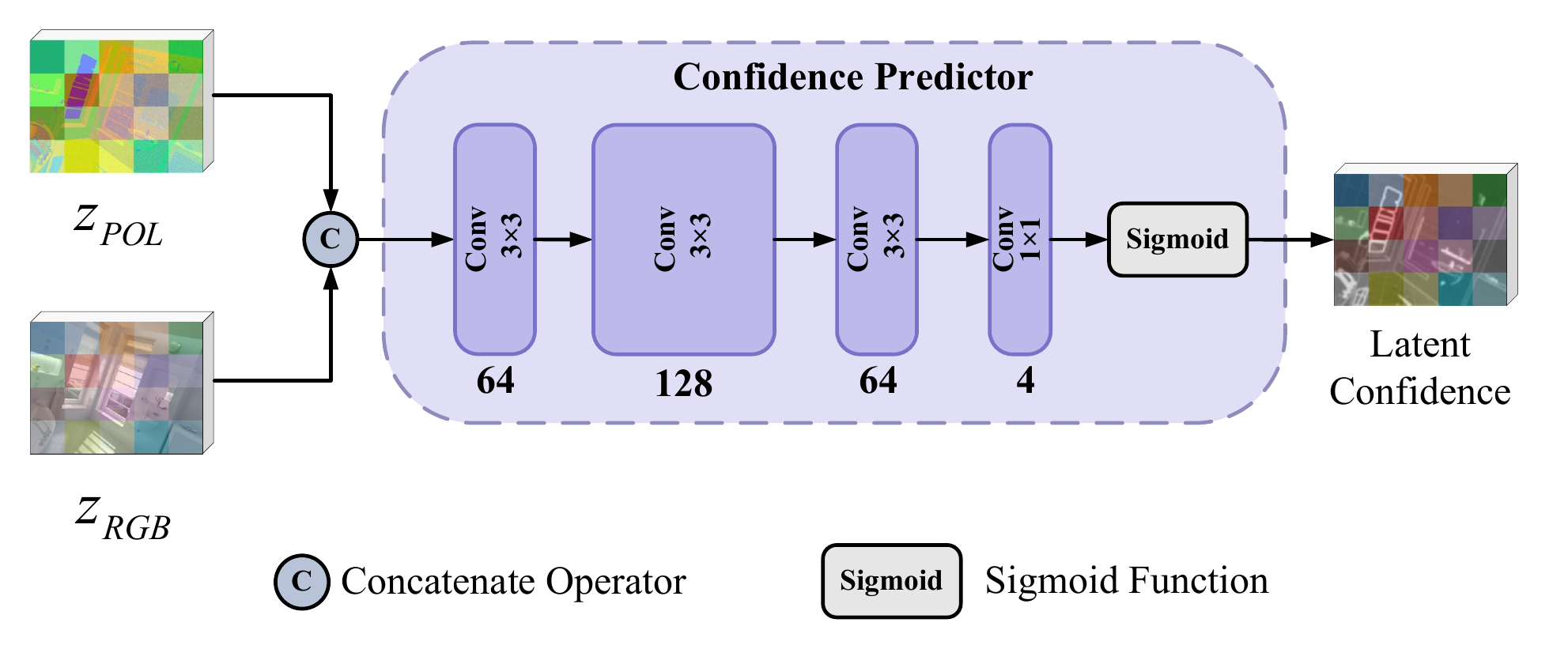}
\caption{Network architecture of the proposed Confidence Predictor. The latent features from the RGB and polarization branches are concatenated and passed through a lightweight convolutional network to produce the confidence map $\alpha$, indicating the reliability of each modality.}
\label{fig_4}
\end{figure}

The network architecture of the confidence predictor shown in Fig.~\ref{fig_4} adopts a lightweight convolutional neural network (CNN), whose total parameter count is negligible compared to that of the backbone network. As shown in Section V-D, this architecture offers a favorable balance between model capacity and training cost, while retaining strong predictive effectiveness.

\subsection{Gated Latent Fusion}
Given the latent representations of the RGB and polarization modalities, denoted as $\boldsymbol{z}_{RGB}$ and $\boldsymbol{z}_{POL}$, along with the corresponding confidence map $\boldsymbol{\alpha}$, we design a gated latent fusion strategy to enable adaptive cross-modal reconstruction. The fusion operation is defined as:

\begin{equation}
\boldsymbol{z}_f = \boldsymbol{\alpha}*\boldsymbol{z}_{POL}+(1-\boldsymbol{\alpha})*\boldsymbol{z}_{RGB},
\end{equation}
where $\boldsymbol{z}_f$ denotes the fused latent representation, which is subsequently used in the diffusion denoising process for depth reconstruction. This formulation performs spatially adaptive weighting of the two modalities at each position in the latent space, allowing the model to selectively emphasize the more reliable source of information depending on local surface characteristics. 

The $\alpha$-map is produced by the learnable confidence predictor mentioned in Section IV-B , which is trained jointly with the main U-Net denoiser. In such task, the quality of $\alpha$-map directly influences the structure of latent representations fed into the denoising U-Net. Conversely, the U-Net’s denoising objective provides essential supervision signals that guide the learning of meaningful confidence maps. This tightly coupled relationship necessitates joint optimization: it allows the confidence predictor to adaptively learn fusion strategies that align with the end-task objective, rather than relying on fixed or heuristic fusion rules. As a result, the network achieves fully trainable, task-specific information integration in a seamless and consistent manner.

\subsection{Depth Estimation in Latent Space}
Our framework is built upon Stable Diffusion V2\cite{rombach2022high}, a pretrained generative model trained on the large-scale LAION-5B dataset\cite{schuhmann2022laion}. This extensive pretraining endows the model with rich image priors, offering valuable support for monocular depth estimation tasks.

In this work, we retain the Marigold\cite{ke2024repurposing} backbone and extend its input interface to include both RGB and polarization images. The inference process consists of three major components: a Variational Autoencoder (VAE), a conditional U-Net, and a noise scheduler. The VAE compresses the input into a latent representation, while the decoder reconstructs the predicted depth map from the final latent variable $\boldsymbol{z}_0$.The U-Net acts as a denoiser in the diffusion framework, conditioned on the fused latent features.

Compared to modeling directly in image space, operating in latent space offers several advantages, including reduced dimensionality and a more regular latent distribution, which lead to improved computational efficiency and training stability.During training, we apply Gaussian noise to the VAE-encoded depth latents to simulate the forward diffusion process:

\begin{equation}
\boldsymbol{z}_t = \sqrt{\bar{\boldsymbol{\alpha}}_t}\boldsymbol{z}_0 + \sqrt{\boldsymbol{1} - \bar{\boldsymbol{\alpha}}_t} \boldsymbol{\varepsilon},
\end{equation}
where $\bar{\boldsymbol{\alpha}}_t$ is the noisy scheduler terms used to control sample quality.

The U-Net denoiser $\mathcal{E}_{\theta}$ is trained to predict the noise component $\boldsymbol{\varepsilon}$  added to the latent variable  $\boldsymbol{z}_t$, enabling conditional denoising process. This follows the standard formulation in diffusion models. The training objective is defined as:

\begin{equation}
\mathcal{L}_{d} = \mathbb{E}_{\boldsymbol{z}_0, \boldsymbol{\varepsilon}, t} \left[ \left\| \mathcal{E}_\theta \left(\boldsymbol{z}_t, t, \boldsymbol{z}_f \right) - \boldsymbol{\varepsilon} \right\|^2 \right].
\end{equation}

Although recent studies have explored single-step diffusion methods\cite{garcia2025fine}\cite{he2024lotus}\cite{xu2024matters} to improve inference efficiency, we adopt the original multi-step diffusion scheme as in Marigold\cite{ke2024repurposing}. This choice is primarily motivated by the limitations of the single-step paradigm, which typically employs a direct regression objective—treating the U-Net output as the final prediction and supervising it against ground-truth depth or normals.

Under this setting, the model is required to simultaneously perform latent feature fusion, semantic understanding, and high-precision reconstruction within a single forward pass. Such tight coupling often compromises the effectiveness of the cross-modal fusion mechanism between RGB and polarization inputs.

This issue is particularly pronounced during the early training stages, when the latent space is not yet robust to polarization noise and the semantic alignment across modalities remains underdeveloped. In these conditions, forcing the model to immediately rely on both modalities for final prediction frequently leads to the collapse of the fusion strategy—manifested as the $\alpha$-map  rapidly converging to zero, effectively disabling the polarization path.

In contrast, multi-step diffusion offers a progressive refinement process of the latent representation, enabling the model to stably learn the complementarity and adaptive weighting between modalities. This leads to more reliable predictions and improved generalization.

Building on these parts, our training pipeline is shown in Fig.~\ref{fig_3}.
\section{Experiment}
\subsection{Experiment Implementation}
\textbf{Training Details}: The entire training pipeline was implemented using PyTorch\cite{paszke2019pytorch} under Ubuntu 22.04. We adopt Stable Diffusion V2 \cite{jung2022my} as the backbone model, retaining its original pretrained weights and objective function. All input images are resized to 640×480, and the model is trained for 20,000 iterations on a single NVIDIA RTX 4090 GPU with a batch size of 32, consuming approximately 3 days. We use the Adam\cite{kingma2014adam} optimizer with an initial learning rate of $3 \times 10^{-5}$ for the U-Net, along with a cosine annealing schedule. The learning rate for the Confidence Predictor is fixed at $1 \times 10^{-4}$.

To provide a comprehensive evaluation of both accuracy and efficiency, we report results under two inference configurations: a high-precision setting that performs 50-step DDIM sampling (denoted as Ours-Standard), and an accelerated setting that adopts a trailing sampling schedule with only 4 denoising steps (denoted as Ours-Accelerated). Both results are obtained from the same model architecture and training procedure; the only difference lies in the diffusion sampling strategy used during inference. The accelerated configuration yields substantially faster runtime while retaining competitive prediction accuracy.

\textbf{Dataset}: Our model is mainly trained on the proposed HyperPol dataset, which provides synthetic polarization observations with physically consistent AoLP/DoLP patterns. In the data synthesis process, we make a necessary simplification in handling the $\pi$ and $\pi/2$ ambiguities inherent in polarization imaging. As discussed in Section IV-A, the adopted polarization encoding scheme effectively mitigates any potential errors introduced by this simplification. The detailed dataset generation pipeline is presented in the Supplemental Material. 

Additionally, we incorporate the real-world multi-modal dataset HAMMER\cite{jung2022my} to enhance model generalization. During training, the model samples data from HyperPol and HAMMER with a ratio of 0.9 and 0.1 respectively.

\textbf{Benchmark Metrics}: We report Absolute Mean Relative Error (AbsRel), $\delta1$ accuracy and $\delta2$ accuracy to evaluate the depth estimation performance. Experiments are conducted on both the HyperPol and HAMMER datasets.

\subsection{Comparison with Other Depth Estimation Methods}
In this section, we compare CDPR with other state-of-the-art monocular depth estimation approaches in both qualitative and quantitative aspects. Since all compared baselines operate on RGB-only inputs, we restrict our evaluation to the RGB modality from each dataset to ensure a fair comparison. In addition, our evaluation is designed under the standard monocular depth estimation setting, and we focus on representative methods with comparable problem formulations.

\begin{figure*}[!t]
\centering
\includegraphics[width=7.2in]{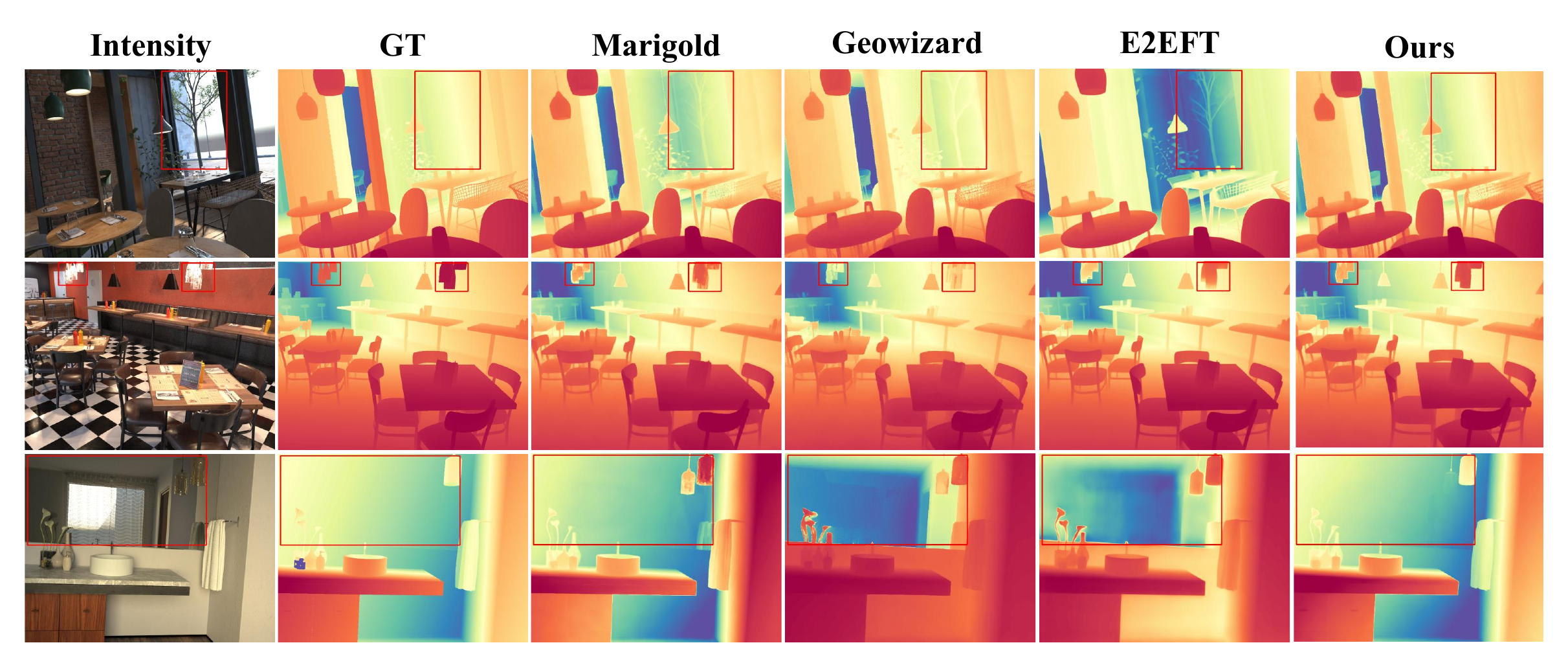}
\caption{Qualitative comparison of depth estimation results on HyperPol dataset. Compared to prior methods, our method produces more accurate and structurally consistent depth maps, particularly in regions with reflective or transparent surfaces (highlighted by red boxes).}
\label{fig_5}
\end{figure*}

\begin{figure*}[!t]
\centering
\includegraphics[width=7.2in]{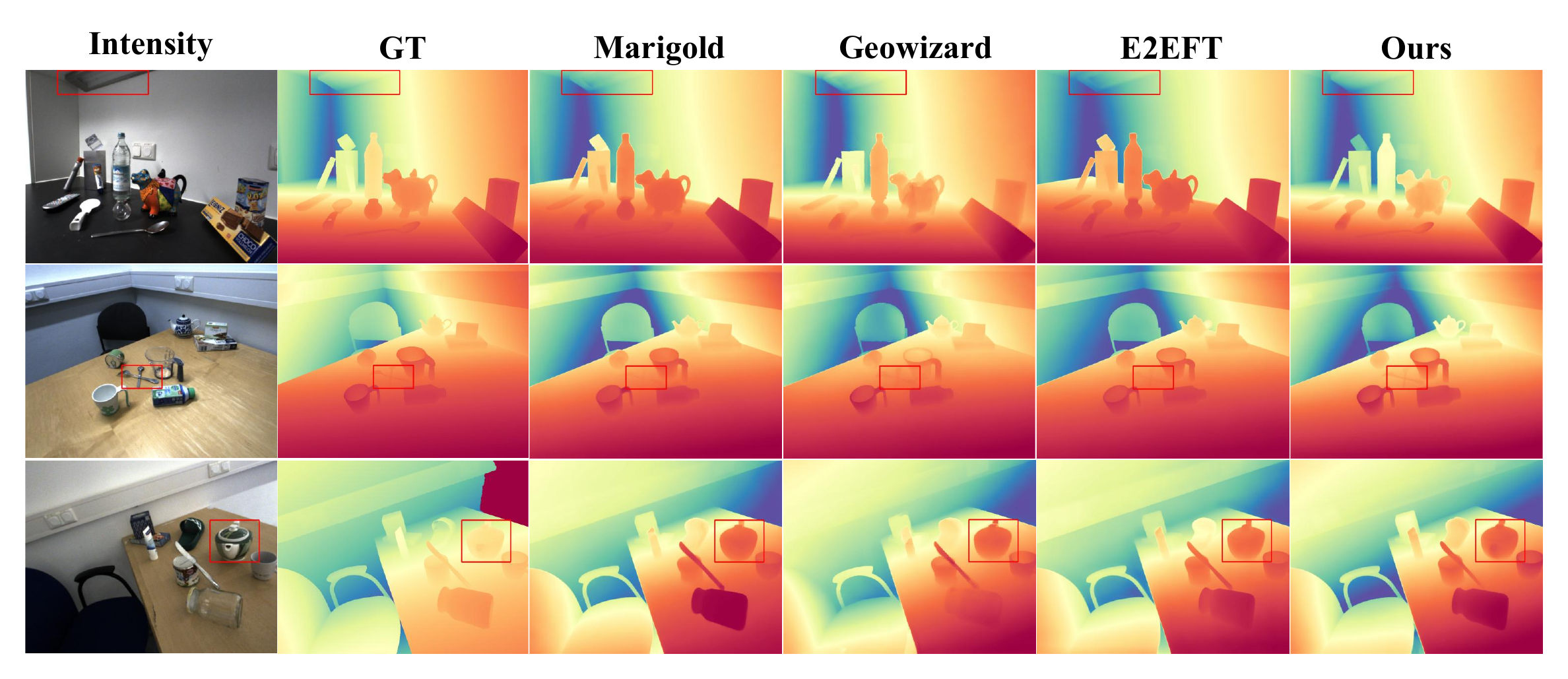}
\caption{Qualitative comparison of depth estimation results on HAMMER dataset. Compared to the more complex HyperPol scenes, the HAMMER dataset features simpler indoor environments with relatively limited depth variation. In such cases, RGB-based methods already produce satisfactory results, leaving less room for improvement. While our method shows less visible enhancement in overall depth quality under these conditions, it still provides noticeable advantages in restoring fine geometric details, such as the teapot spout in the third example.}
\label{fig_6}
\end{figure*}

\begin{figure*}[!t]
\centering
\includegraphics[width=7.2in]{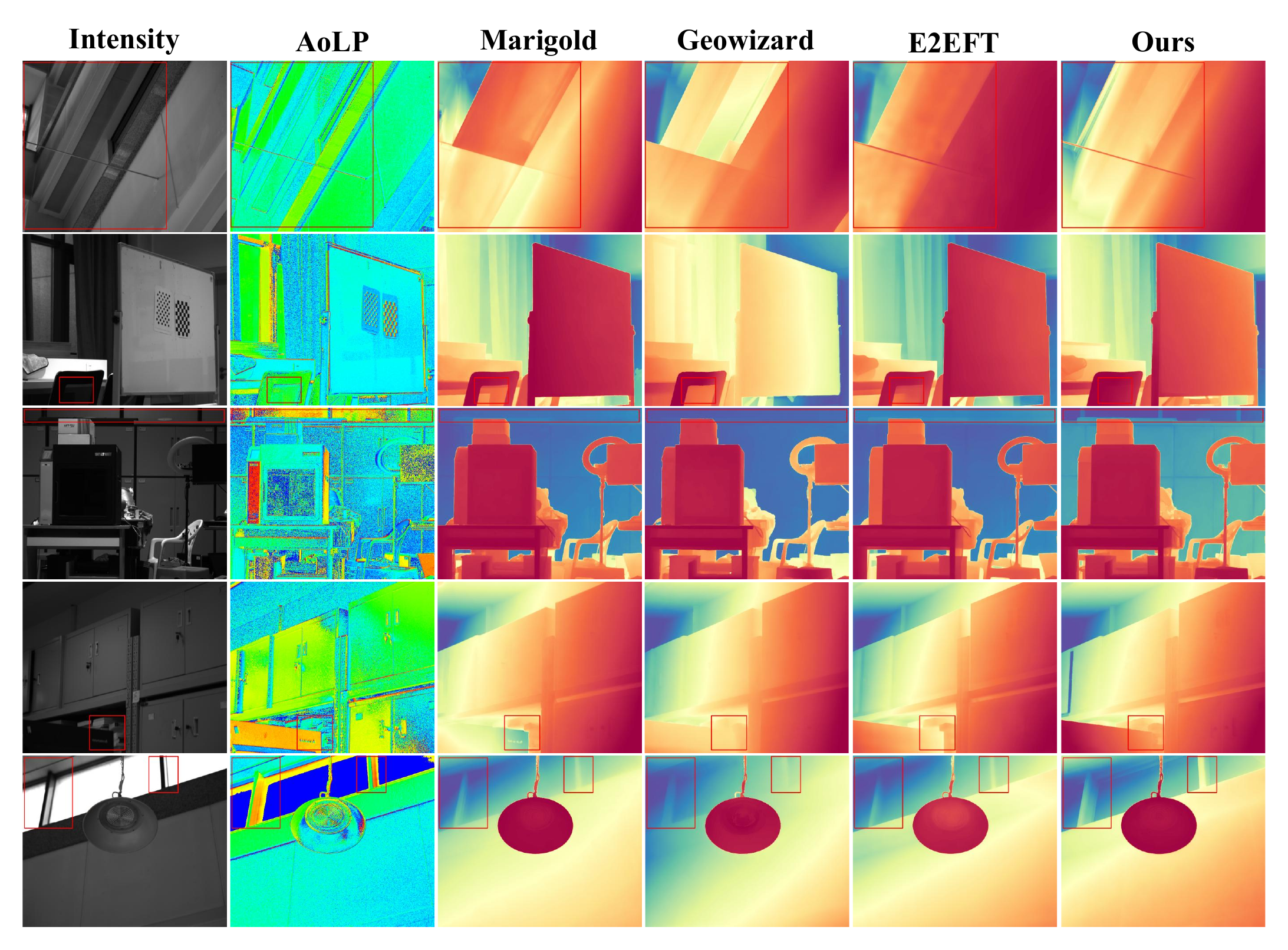}
\caption{Qualitative comparison of depth estimation results on in-the-wild examples. As highlighted in the red boxes, RGB-only methods tend to produce ambiguous depth estimates in challenging regions, whereas incorporating polarization cues leads to more coherent depth structures and improved boundary consistency.}
\label{fig_inthewild}
\end{figure*}

\begin{table*}[ht]
    \centering
    \begin{minipage}{0.65\textwidth}
        \centering
        \caption{Quantitative Result of Monocular Depth Estimation}
        \label{tab:chap:table_depth}
        \begin{tabular}{lccc|ccc}
            \toprule
            \textbf{Method} & \multicolumn{3}{c|}{\textbf{HyperPol}} & \multicolumn{3}{c}{\textbf{HAMMER}} \\
            & \textbf{AbsRel ↓} & \boldmath{$\delta_1$ ↑} & \boldmath{$\delta_2$ ↑} & \textbf{AbsRel ↓} & \boldmath{$\delta_1$ ↑} & \boldmath{$\delta_2$ ↑} \\
            \midrule
            DPT\cite{ranftl2021vision} & 24.2 & 68.1 & 87.8 & 7.6 & 95.5 & 99.5 \\
            MiDaS\cite{ranftl2020towards} & 22.4 & 71.2 & 89.4 & 8.2 & 94.6 & 99.4 \\
            Omnidata\cite{eftekhar2021omnidata} & 18.8 & 77.9 & 92.0 & 8.2 & 92.8 & 99.2 \\
            Marigold v1.0\cite{ke2024repurposing} & 10.5 & 89.7 & 97.4 & 5.3 & 98.6 & 99.8 \\
            Marigold v1.1\cite{ke2025marigold} & 11.2 & 88.7 & 97.1 & 6.4 & 97.5 & 99.7 \\
            Geowizard\cite{fu2024geowizard} & 11.2 & 89.1 & 96.6 & \textbf{5.2} & 99.1 & 99.9 \\
            DepthFM\cite{gui2025depthfm} & 14.8 & 83.0 & 95.0 & 10.8 & 89.6 & 99.1 \\
            Lotus\cite{he2024lotus} & 11.6 & 89.6 & 96.6 & 5.6 & 98.2 & 99.9 \\
            E2EFT-Marigold\cite{garcia2025fine} & 9.3 & 91.6 & 98.0 & 5.4 & 98.1 & 99.8 \\
            E2EFT-SD2\cite{garcia2025fine} & 9.7 & 91.9 & 98.0 & 6.1 & 98.7 & 99.8 \\
            \textbf{Ours-Standard(50 steps)} & \textbf{8.8} & \textbf{93.2} & \textbf{98.5} & 5.3 & \textbf{99.4} & \textbf{99.9} \\
            \textbf{Ours-Accelerated(4 steps)} & 9.6 & 91.9 & 98.0 & 5.7 & 99.0 & 99.9 \\
            \bottomrule
        \end{tabular}  
        \vspace{0.3em}
        
        \captionsetup{justification=raggedright, singlelinecheck=false}
        \caption*{\footnotesize
        \textit{Ours-Standard} uses a 50-step DDIM sampling schedule for maximal accuracy.\\
        \textit{Ours-Accelerated} adopts a trailing sampling schedule with 4 denoising steps for efficient inference.
}

    \end{minipage}
\end{table*}

Our comparative experiments are mainly conducted on 500 test samples from the HyperPol dataset and 200 test samples from the HAMMER dataset. As shown in Table~\ref{tab:chap:table_depth}, our method achieves satisfactory performance across all evaluation metrics on both the HyperPol and HAMMER datasets. Fig. \ref{fig_5} and Fig. \ref{fig_6} present qualitative comparisons against other generative baseline methods. Our approach demonstrates superior structural consistency in challenging regions, such as specular surfaces, transparent materials, and textureless areas—scenarios where RGB-only methods typically struggle to produce coherent depth predictions.

While the improvements are clearly visible on the complex HyperPol scenes, the performance gains on the HAMMER dataset appear less prominent. This is mainly due to the simpler indoor environments and limited depth variation in HAMMER, where RGB-based methods already achieve strong baseline results. Nonetheless, our method still offers consistent advantages in restoring fine geometric details—such as the teapot spout in the third example of Fig. \ref{fig_6} and effectively alleviates the scale ambiguity issues discussed in Fig.\ref{fig_1}, demonstrating the robustness and reliability of polarization guidance even in less demanding scenarios.

In addition, to further assess robustness and generalization in open and real-world inference scenarios, we also collect a set of in-the-wild examples using a real polarization camera \emph{Daheng MER2-503-23GC-P-POL}, which simultaneously acquires four polarization-aligned intensity images in a single shot. Although these samples do not provide ground-truth depth annotations, they enable qualitative evaluation under uncontrolled conditions with real sensor noise, complex materials, and diverse lighting, offering complementary evidence of the practical robustness of our method. Qualitative comparison of these samples are presented in Fig. \ref{fig_inthewild}. The results indicate that our method maintains stable depth predictions across a variety of real-world scenarios.

Furthermore, a constrained evaluation on the HouseCat6D\cite{jung2024housecat6d} dataset using object-centric depth annotations is provided in the supplementary material, offering complementary insights into method behavior on real sensor data.

\subsection{Computational Efficiency Analysis}
To analyze the computational characteristics of our method, we evaluate the trade-off between inference efficiency and prediction accuracy on the HyperPol dataset, comparing our approach with baseline methods under different sampling configurations. Our primary setting (Ours-Standard) adopts the original multi-step DDIM sampler used in Marigold v1.0\cite{ke2024repurposing}, while we additionally evaluate an accelerated inference variant inspired by the Marigold v1.1\cite{ke2025marigold} framework. All efficiency evaluations are conducted on a single NVIDIA RTX 4090 GPU using input images with a resolution of 1024 × 768.

Specifically, we adopt the trailing timestep schedule and enable the zero-SNR configuration during fine-tuning, following the inference strategy of Marigold v1.1. Unlike the full v1.1 training recipe, which further incorporates data augmentation techniques such as flipping, blurring, and color jitter, our implementation keeps the training data pipeline consistent with v1.0. This controlled setup allows us to isolate and assess the impact of the modified sampling schedule on both inference efficiency and prediction accuracy.

It is worth noting that the proposed accelerated inference variant (4 steps) and the standard variant (50 steps) share exactly the same multi-step diffusion training objective and model architecture, and differ only in the inference-time sampling schedule. As such, they represent an efficiency–accuracy trade-off within the same diffusion paradigm, and are not affected by the limitations associated with single-step regression-based formulations discussed in Section IV-D.

Table~\ref{tab:tradeoff_table} summarizes the results across different sampling strategies. As expected, reducing the number of denoising steps leads to a substantial improvement in inference speed at the cost of a moderate accuracy drop. Importantly, CDPR retains strong performance in terms of accuracy, even under the accelerated configuration, indicating that the proposed confidence-guided fusion preserves most of the predictive capability of the full multi-step diffusion process.

Overall, these results highlight a clear efficiency–accuracy continuum within our framework: the standard multi-step configuration is preferable for accuracy-critical depth reconstruction, whereas the accelerated trailing-based variant offers a practical alternative for scenarios where inference efficiency and real-time performance are prioritized.

\begin{table}[t]
    \centering
    \caption{Inference Time of CDPR and Other Methods}
    \label{tab:tradeoff_table}
    \begin{tabular}{lcc}
        \toprule
        \textbf{Method} & \textbf{Time(s) / img} & \textbf{Inference step} \\
        \midrule
        DPT & 0.21 & -- \\
        Marigold v1.0 & 3.77 & 50 \\
        Marigold v1.1 & 0.75 & 4 \\
        Geowizard & 2.68 & 10 \\
        E2EFT-Marigold & 0.56 & 1 \\
        E2EFT-SD2 & 0.59 & 1 \\
        \textbf{Ours-Standard (50 steps)} & 3.81 & 50 \\
        \textbf{Ours-Accelerated(4 steps)} & 0.81 & 4 \\
        \bottomrule
    \end{tabular}
\end{table}

\begin{figure*}[!t]
\centering
\includegraphics[width=6.5in]{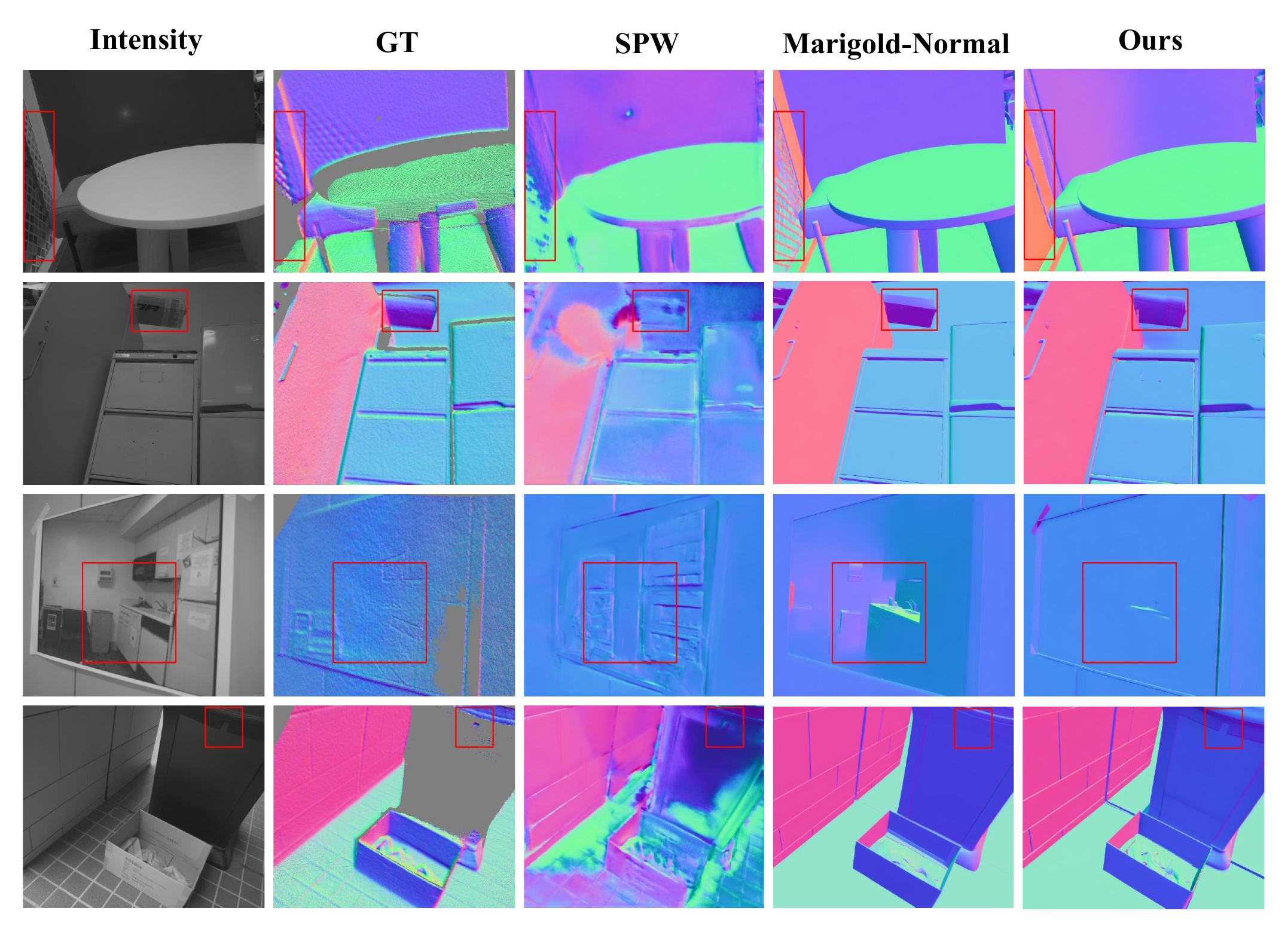}
\caption{Qualitative results of our extended experiments on surface normal estimation. The results show that our method remains effective for normal prediction, outperforming existing baselines. Notably, our model suppresses pseudo-textures caused by RGB-only inputs and yields cleaner, more faithful geometry in challenging regions (highlighted by red boxes).}
\label{fig_7}
\end{figure*}

\begin{table*}[ht]
    \centering
    \caption{Quantitative Result of Surface Normal Estimation}
    \label{tab:chap:table_normal}
    \begin{tabular}{lccc|ccc}
        \toprule
        \textbf{Method} & \multicolumn{3}{c|}{\textbf{Angular Error} ↓} & \multicolumn{3}{c}{\textbf{Accuracy} ↑} \\
        & \textbf{Mean} & \textbf{Median} & \textbf{RMSE} & \boldmath{$11.25^\circ$} & \boldmath{$22.5^\circ$} & \boldmath{$30.0^\circ$} \\
        \midrule
        SPW\cite{lei2022shape} & 17.60 & 13.85 & 22.57 & 46.7 & 77.3 & 85.5 \\
        Marigold-Normal\cite{ke2025marigold} & 12.76 & 9.74 & 17.64 & 71.3 & 86.0 & 89.8 \\
        \textbf{Ours} & \textbf{12.23} & \textbf{8.98} & \textbf{17.21} & \textbf{73.5} & \textbf{87.3} & \textbf{90.4} \\
        \bottomrule
    \end{tabular}
\end{table*}

\subsection{CDPR for Surface Normal Estimation}
To further validate the generalizability and scalability of our CDPR framework, we extend the task to surface normal estimation. In this setting, we retain the same architectural design and fusion mechanism, adapting only the output target to predict per-pixel surface normals.

We conduct training solely on HyperPol dataset due to the lack of ground-truth normal annotations in HAMMER dataset. Importantly, since the HyperPol dataset is generated with ground-truth normals from the original Hypersim geometry, we do not use it for quantitative testing to avoid evaluation bias.Instead, we evaluate our model on the publicly available SPW dataset\cite{lei2022shape}, a real-world polarization dataset designed for scene-level surface normal estimation. This dataset does not overlap with any training data and serves as a strong benchmark to assess generalization in real-world settings.

As shown in Table~\ref{tab:chap:table_normal} and Fig. ~\ref{fig_7}, our method demonstrates strong performance on the SPW dataset. Beyond validating the scalability of the CDPR framework for diffusion-based dense prediction, the results also highlight the advantage of our synthetic polarization data in guiding the model toward more reliable and generalizable predictions. Notably, our approach effectively suppresses texture-copying artifacts commonly observed in RGB-based methods, and achieves superior recovery of fine structural details, particularly in regions with subtle geometry or reflective surfaces.

\subsection{Ablation Study}
To better evaluate the contribution of each module, we conduct comprehensive ablation experiments on the HyperPol dataset. We continue to use AbsRel, $\delta1$ accuracy and $\delta2$ accuracy as the primary evaluation metrics.

\textbf{1.RGB and Polarization Inputs}: We begin by examining the necessity of incorporating both RGB and polarization inputs. To ensure fair comparison, we replace the multimodal inputs with same-modality dual-channel combinations while keeping all other network components.

Specifically, we construct two ablation variants: one that replaces the original RGB + POL input with two RGB images (RGB + RGB), and another using two polarization images (POL + POL). These settings allow us to quantitatively assess the contribution of each modality.

\begin{figure}[!t]
\centering
\includegraphics[width=3.5in]{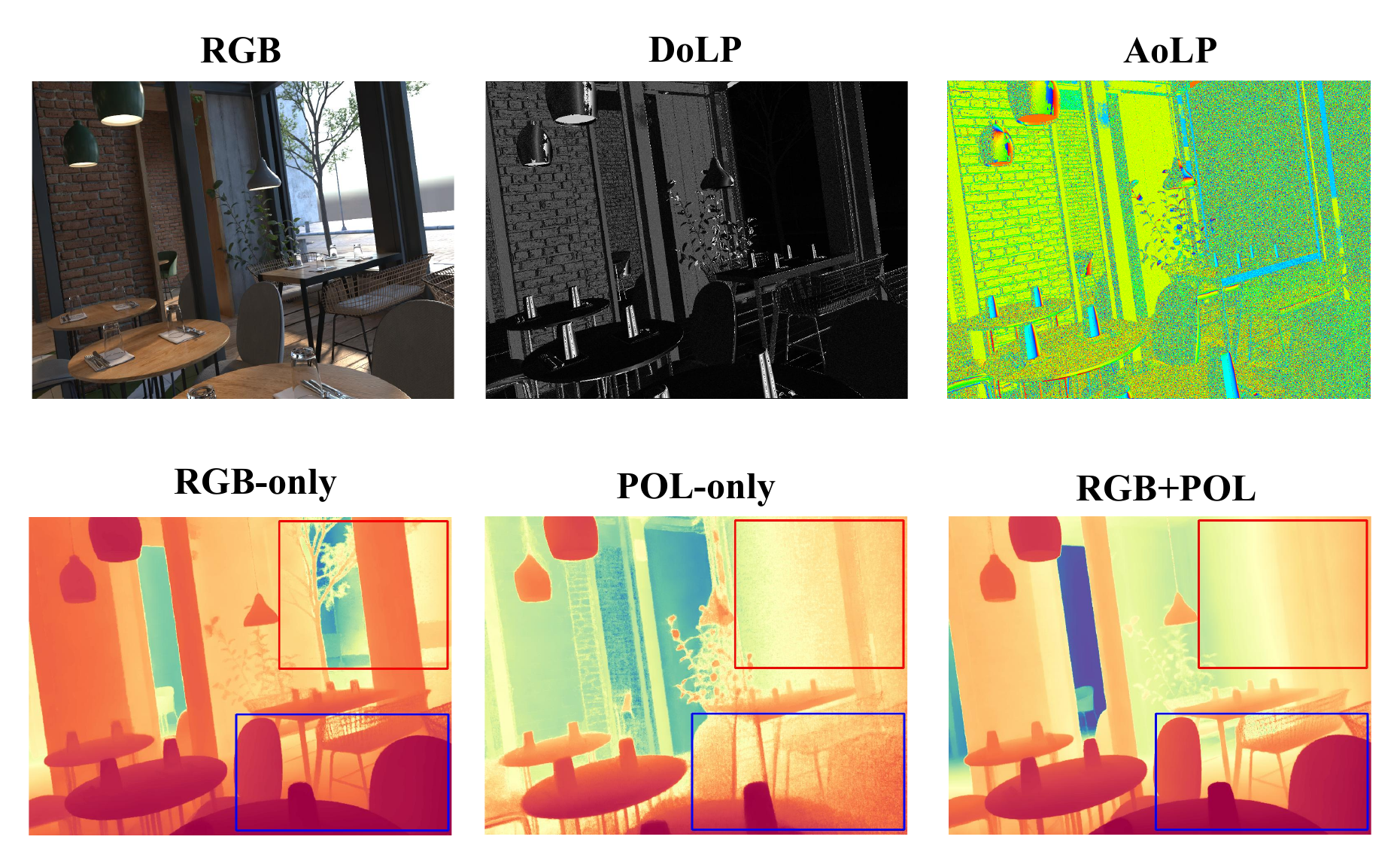}
\caption{Comparison of RGB input, polarization cues, and depth predictions obtained under different modality configurations. The top row shows the input RGB image, DoLP, and AoLP. The bottom row presents depth predictions using RGB-only, POL-only, and our RGB+POL adaptive fusion guided by the confidence predictor. Regions where RGB provides more reliable geometric cues are highlighted with blue box, while regions where polarization cues are more advantageous are highlighted with red box. Our confidence-guided fusion strategy effectively combines the strengths of both modalities, resulting in more reliable and accurate depth estimation.}
\label{add_fig1}
\end{figure}

As shown in Table~\ref{tab:chap:table_1}, both unimodal combinations (RGB-only or POL-only) lead to significant performance degradation in both accuracy and error metrics compared to the original RGB + POL configuration. The result indicates that RGB and polarization images provide complementary cues in the latent space, and their joint usage plays a crucial role in enhancing the model’s depth estimation performance.

As further illustrated in Figure~\ref{add_fig1}, different modalities exhibit distinct strengths and weaknesses in challenging areas. RGB-only predictions suffer from ambiguity in reflective or transparent regions, such as glass surfaces, where appearance cues become unreliable (highlighted with red box), while POL-only predictions more clearly delineate these challenging areas, but they exhibits degraded global performance due to noise and instability in polarization measurements(highlighted with blue box). In contrast, our confidence-guided fusion strategy (RGB+POL) adaptively integrates complementary strengths from both modalities, leading a more reliable depth estimation result.

\begin{table}[ht]
    \centering
    \caption{Ablation of Input}
    \label{tab:chap:table_1}
    \begin{tabular}{>{\raggedright\arraybackslash}p{1.5cm} ccc}
        \toprule
        {} & \textbf{AbsRel ↓} & \boldmath{$\delta_1$ ↑} & \boldmath{$\delta_2$ ↑} \\
        \midrule
        w/o RGB & 18.5 & 78.3 & 92.1 \\
        w/o POL & 12.9 & 86.8 & 96.0 \\
        \textbf{Full Input} & \textbf{8.8} & \textbf{93.2} & \textbf{98.5} \\
        \bottomrule
    \end{tabular}
\end{table}

\textbf{2.Effectiveness and Noise Robustness of Different Fusion strategies}: To evaluate the effectiveness of our proposed confidence predictor in cross-modal fusion, we compare four fusion strategies under identical network architecture and training hyperparameters. In addition to clean inputs, we systematically inject Gaussian noise with increasing intensity into the polarization branch at the stage after the polar encoder and before the VAE encoder to examine the robustness of each fusion method, as shown in Table~\ref{tab:chap:table_2}. The injection point is chosen to emulate residual noise patterns that the polar encoder cannot entirely remove in practical scenarios, thereby providing a more realistic stress test for the fusion mechanism.

The compared fusion strategies include:

\noindent
$\bullet$ \textbf{Early Fusion.} RGB and polarization images are element-wise added at the input level before being passed into the VAE encoder, with no confidence predictor or explicit fusion module involved.

\noindent
$\bullet$ \textbf{Fixed Fusion.} Latent features from the RGB and polarization branches are fused via pixel-wise averaging without any explicit confidence estimation.

\noindent
$\bullet$ \textbf{Random Fusion.} The fusion coefficient $\alpha$ is randomly sampled from a uniform distribution in $[0,1]$, simulating an unconstrained gating mechanism. The experiment is repeated 10 times, and the reported result is the average value across these runs.

Real-world polarization noise is inherently complex and exhibits spatially coupled disturbances. These disturbances arise from multiple factors, including sensor characteristics, illumination conditions, and material properties, making them difficult to accurately model in a controlled and reproducible manner. 

Nevertheless, injecting progressively stronger global Gaussian noise provides a tractable and consistent approximation, enabling systematic evaluation of the robustness of each fusion strategy under degraded polarization conditions. Concretely, given the normalized polarization representation $\boldsymbol{x}_p$, we perturb it by

\begin{equation}
\widetilde{\boldsymbol{x}}_p = 
\mathrm{clip}\!\left(
\boldsymbol{x}_p + \beta \cdot \mathcal{N}(0, \mathbf{I}),
\, -1,\, 1
\right),
\label{eq:noise_injection}
\end{equation}

where $\beta$ denotes the noise intensity and clipping ensures the perturbed signal remains within the valid range. To more clearly quantify the robustness of each method, we additionally report in the last row of Table~\ref{tab:chap:table_2} the performance degradation between the zero-noise and highest-noise intensities ($\beta$ = 0 vs.\ $\beta$ = 1.0), which serves as a compact indicator of stability. 

Across all noise intensities, our confidence-guided fusion achieves the best performance in all metrics, showing both higher accuracy and stronger robustness. While Early Fusion and Random Fusion degrade rapidly as noise increases, and Fixed Fusion shows moderate sensitivity, our method maintains consistently superior depth estimation even when the polarization signal becomes heavily corrupted. Notably, the moderate stability observed in the Fixed Fusion baseline also suggests that the diffusion-based backbone provides inherent resilience to noise when the input remains reasonably stable. On top of this backbone robustness, our learned confidence map further enhances stability by adaptively modulating the contribution of each modality.


\begin{table*}[ht]
    \centering
    \caption{Quantitative Result of Four Fusion Strategies Under Different Noise Intensities}
    \label{tab:chap:table_2}
    \begin{tabular}{l  ccc|ccc|ccc|ccc}
        \toprule
        \textbf{\textbf{Noise Intensity $\beta$}} 
        & \multicolumn{3}{c|}{\textbf{Early Fusion}} 
        & \multicolumn{3}{c|}{\textbf{Fixed Fusion}} 
        & \multicolumn{3}{c|}{\textbf{Random fusion}} 
        & \multicolumn{3}{c}{\textbf{Ours Fusion}} \\
        
        & \textbf{AbsRel ↓} & \boldmath{$\delta_1$ ↑} & \boldmath{$\delta_2$ ↑} 
        & \textbf{AbsRel ↓} & \boldmath{$\delta_1$ ↑} & \boldmath{$\delta_2$ ↑}
        & \textbf{AbsRel ↓} & \boldmath{$\delta_1$ ↑} & \boldmath{$\delta_2$ ↑}
        & \textbf{AbsRel ↓} & \boldmath{$\delta_1$ ↑} & \boldmath{$\delta_2$ ↑} \\
        \midrule
        
        0
        & 13.2 & 85.9 & 95.6  
        & 9.7  & 91.7 & 97.8
        & 13.7   & 85.3   & 95.7 
        & 8.8   & 93.2   & 98.4 \\
        
        0.1 
        & 13.2 & 85.8 & 95.5  
        & 9.7 & 91.7 & 97.8
        & 13.7   & 85.3   & 95.6 
        & 8.9   & 92.9   & 98.4 \\
        
        0.3 
        & 13.4 & 85.5 & 95.5  
        & 9.8 & 91.4 & 97.7
        & 13.8   & 85.1   & 95.6 
        & 9.1   & 92.8   & 98.2 \\
        
        0.5
        & 13.4 & 85.4 & 95.5  
        & 10.0  & 91.1 & 97.5
        & 14.1   & 84.6   & 95.5
        & 9.1   & 92.7   & 98.2 \\
        
        0.7 
        & 14.6 & 83.3 & 94.7  
        & 10.3 & 90.7 & 97.2
        & 14.4   & 83.9   & 95.2 
        & 9.2   & 92.5   & 98.2 \\
        
        1.0 
        & 17.5 & 79.6 & 93.0  
        & 10.7 & 90.2 & 96.9
        & 15.4   & 82.2   & 94.4 
        & 9.5   & 92.2   & 98.0 \\

        \midrule
        \textbf{Degradation} ↓
        & 4.3 & 6.3 & 2.6
        & 1.0 & 1.5 & 0.9
        & 1.7 & 3.1 & 1.3
        & \textbf{0.7} & \textbf{1.0} & \textbf{0.4}\\
        
        \bottomrule
    \end{tabular}
\end{table*}

\textbf{3.Backbone of Confidence Predictor}: To evaluate how the backbone architecture of the confidence predictor affects the fusion performance, we perform a comparative study using four different lightweight network designs under identical training settings. This comparison is motivated by our goal of maintaining a lightweight and efficient confidence predictor without compromising performance.

\noindent
$\bullet$ \textbf{Simple MLP}: A fully connected network that maps latent features to the \textbf{$\alpha$}-map, suitable for processing vectorized representations.

\noindent
$\bullet$ \textbf{ResNet-18}\cite{he2016deep}: A deeper residual network designed to facilitate gradient flow through skip connections.

\noindent
$\bullet$ \textbf{Standard U-Net}\cite{ronneberger2015u}: A standard encoder-decoder structure with symmetric skip connections, capable of multi-scale feature aggregation.

\noindent
$\bullet$ \textbf{Transformer (Light)}\cite{vaswani2017attention}: A lightweight Transformer-based network consisting of a shallow self-attention encoder with patch embedding and a small number of layers, designed to capture global contextual information with limited computational overhead.

\noindent
$\bullet$ \textbf{Simple CNN(Ours Architecture)}: A lightweight network composed of two convolutional layers, featuring a small receptive field and minimal parameter count.

Surprisingly, the Simple CNN backbone of Figure \ref{fig_4} outperformed all other architectures across multiple evaluation metrics, while also having almost the fewest parameters. This observation aligns with our empirical finding that the $\alpha$-maps produced by Simple CNN converge faster and exhibit greater spatial stability.

In contrast, although U-Net and ResNet-18 offer higher representational capacity, they tend to overfit or behave unstably in this task. We attribute this to the inherently low-dimensional latent space and the lightweight nature of confidence prediction, which does not necessitate deep or complex architectures. The MLP model, lacking spatial locality and inductive bias, performs worst due to its inability to capture structured patterns for spatial fusion. Similarly, the Transformer-based network do not show clear performance advantages in this task, suggesting that their higher modeling flexibility does not translate into improved confidence prediction under the lightweight and low-dimensional latent setting considered here.

These results in Table~\ref{tab:chap:table_3} suggest that confidence prediction is essentially a \textbf{local guidance problem}—the fusion weight $\alpha$ at each spatial location primarily depends on the surrounding latent context rather than on global semantics. Therefore, a shallow CNN backbone is sufficient to learn effective fusion weights in latent space, without introducing excessive model complexity or training instability.

\begin{table}[ht]
	\centering
	\caption {Ablation of Confidence Predictor's Backbone}
	\label{tab:chap:table_3}
	\begin{tabular}{>{\raggedright\arraybackslash}p{2cm} cccc}
		\toprule
		{\textbf{Backbone}} & \textbf{AbsRel ↓} & \boldmath{$\delta_1$ ↑} & \boldmath{$\delta_2$ ↑} & \textbf{Parameters} ↓ \\
		\midrule
		Simple MLP & 9.6 & 91.8 & 97.9 & \textbf{Less than 10K}\\
        ResNet-18 & 9.1 & 92.6 & 98.2 & 13M\\
		Standard U-Net & 9.0 & 93.0 & 98.3 & 30M\\
        Transformer(Light) & 9.2 & 92.7 & 98.2 & 450K\\
		\textbf{Simple CNN} & \textbf{8.8} & \textbf{93.2} & \textbf{98.5} & 100K\\
		\bottomrule
	\end{tabular}
\end{table}

\section{Conclusion}
In this paper, we propose CDPR, a cross-modal diffusion framework for monocular depth estimation that leverages both RGB and polarization cues. By introducing a confidence-aware gated fusion mechanism in the latent space, our model adaptively integrates complementary visual information, effectively mitigating the limitations of RGB-only depth inference. Additionally, we construct HyperPol, a synthetic dataset with rendered polarization annotations, to facilitate the training and evaluation of polarization-guided models. Beyond depth estimation, we further demonstrate that CDPR can be extended to polarization-guided surface normal estimation, showing strong adaptability across dense prediction tasks. Through extensive experiments on both synthetic and real-world datasets, we validate that CDPR excels in challenging regions, while remaining competitive in general scenes. Notably, our work highlights the potential of polarization as a valuable geometric cue and presents a principled, scalable approach for cross-modal integration within generative dense prediction frameworks.

While CDPR achieves strong performance across both synthetic and real datasets, several engineering aspects of the framework present opportunities for further refinement, particularly in improving robustness under real-world polarization noise. The proposed confidence-based fusion strategy emphasizes the lightweight design of the confidence predictor and the interpretability of the fusion process, while future work may explore more flexible fusion mechanisms. Moreover, our current design avoids introducing polarization as an explicit condition, as this would require substantial architectural changes and may disrupt the pretrained priors of Stable Diffusion v2. Nevertheless, conditional diffusion remains a promising future direction as more suitable conditioning mechanisms for geometric cues become available. Lastly, extending CDPR to more comprehensive geometric tasks, such as joint depth–normal estimation, reflectance recovery represents a promising direction. We believe this opens a new direction toward unifying physics-driven cues and generative priors for robust, single-view 3D understanding, and provides a foundation for extending polarization-guided modeling to more general geometry-oriented frameworks.

\bibliographystyle{IEEEtran}
\bibliography{ref}

\end{document}